\documentclass[lettersize,journal]{IEEEtran}
\usepackage{amsmath,amsfonts}
\usepackage{algorithmic}
\usepackage{algorithm}
\usepackage{array}
\usepackage[caption=false,font=normalsize,labelfont=sf,textfont=sf]{subfig}
\usepackage{textcomp}
\usepackage{stfloats}
\usepackage{url}
\usepackage{pifont}
\usepackage{verbatim}
\usepackage{graphicx}
\usepackage[colorlinks=true, linkcolor=black, citecolor=black]{hyperref}
\usepackage{cleveref}
\usepackage{cite}
\usepackage{amsthm}
\usepackage{amsmath}
\usepackage{multirow}
\usepackage{mathtools}
\usepackage[table]{xcolor}
\usepackage{color}
\usepackage{booktabs}
\begin{document}

\title{Leveraging Text-to-Image Diffusion Models for Unsupervised Visual Object Tracking}

\author{Zhengbo Zhang, \IEEEmembership{Member, IEEE}, Zhigang Tu, \IEEEmembership{Senior Member, IEEE},  Junsong Yuan, \IEEEmembership{Fellow, IEEE}, De Wen Soh, \\ Bo Du, \IEEEmembership{Senior Member, IEEE}%
\thanks{Corresponding author: Zhigang Tu (e-mail: tuzhigang@whu.edu.cn)}
\thanks{Zhengbo Zhang and De Wen Soh are with Information Systems Technology and Design Pillar, Singapore University of Technology and Design, Singapore.}%
\thanks{Zhigang Tu is with the State Key Laboratory of Information Engineering in Surveying, Mapping and Remote Sensing, Wuhan University, China.}%
\thanks{Junsong Yuan is with the Department of Computer Science and Engineering, University at Buffalo, State University of New York, USA.}%
\thanks{Bo Du is with the School of Computer Science, Wuhan University, China.}
}

\newcommand{\zb}[1]{\textcolor[rgb]{0.78, 0.2274, 0.333}{\textbf{#1}}}
\definecolor{yellow}{rgb}{1, 1, 0.7}
\definecolor{Second}{rgb}{1, 0.85, 0.7}
\definecolor{Best}{rgb}{1, 0.7, 0.7}

\maketitle

\begin{abstract}
Unsupervised visual object tracking is a challenging task that requires following arbitrary targets in videos without training on ground-truth annotations. Despite considerable progress, existing state-of-the-art unsupervised trackers often struggle in scenarios that demand fine-grained understanding of semantic and visual structural information within video frames. 
Text-to-image diffusion models are well known for their ability to generate images that accurately reflect the semantics and structures described in the input prompt, demonstrating a strong grasp of visual semantics and structures. Building on this capability, we approach the unsupervised tracking from a new perspective by exploiting the rich semantic knowledge encoded in pretrained text-to-image diffusion models.
To adapt the diffusion models, which are originally developed for image generation, to the tracking task, we reinterpret the models as a bridge between text and image modalities. This connection is realized through the cross-attention mechanism: when both text and an image are input into the models, they highlight the regions of the image that are semantically aligned with the text in the cross-attention maps.
 We therefore learn a prompt that represents the tracking target and activates its corresponding region in the cross-attention map for each frame, which enables object tracking with the diffusion model.
Specifically, our method Diff-Tracking is composed of two main components: an initial prompt learner and an online prompt updater.
 The initial prompt learner generates a prompt that captures the target object in the first frame, allowing the diffusion model to identify the target. The online prompt updater refines the prompt based on motion information, enabling consistent tracking across video frames.
We evaluate our approach on six challenging tracking datasets, showing that Diff-Tracking achieves strong performance compared to existing unsupervised trackers.
\end{abstract}


\section{Introduction}

\IEEEPARstart{V}{isual} object tracking~\cite{wang2025omnitracker} is a fundamental task in computer vision, with broad applications in domains such as autonomous driving~\cite{chen2021novel,gao2019manifold,zhuang20234d} and robotics~\cite{papanikolopoulos1993visual,budiharto2020design,wang2025gv}. Recent advances in the visual object tracking have been driven by deep learning based trackers~\cite{yu2020deformable,danelljan2020probabilistic,foo2023unified}, which have become the dominant paradigm. However, these approaches heavily depend on large amounts of annotated data for supervised training. Such substantial cost and time required for manual annotation have spurred growing interest in unsupervised visual tracking~\cite{zheng2021learning,wang2019unsupervised,shen2022unsupervised,wang2021unsupervised}. 
Despite significant progress in unsupervised methods, with some approaching the performance of supervised counterparts~\cite{zheng2021learning,shen2022unsupervised,wang2021unsupervised}, unsupervised object tracking remains highly challenging. The main difficulties arise in scenarios that demand effective utilization of rich semantic and structural cues within frames~\cite{wang2021unsupervised,sio2020s2siamfc} and the abundant contextual relationships across video sequences~\cite{zheng2021learning}.

\begin{figure}[!t]
\centering
\includegraphics[width=3.5in]{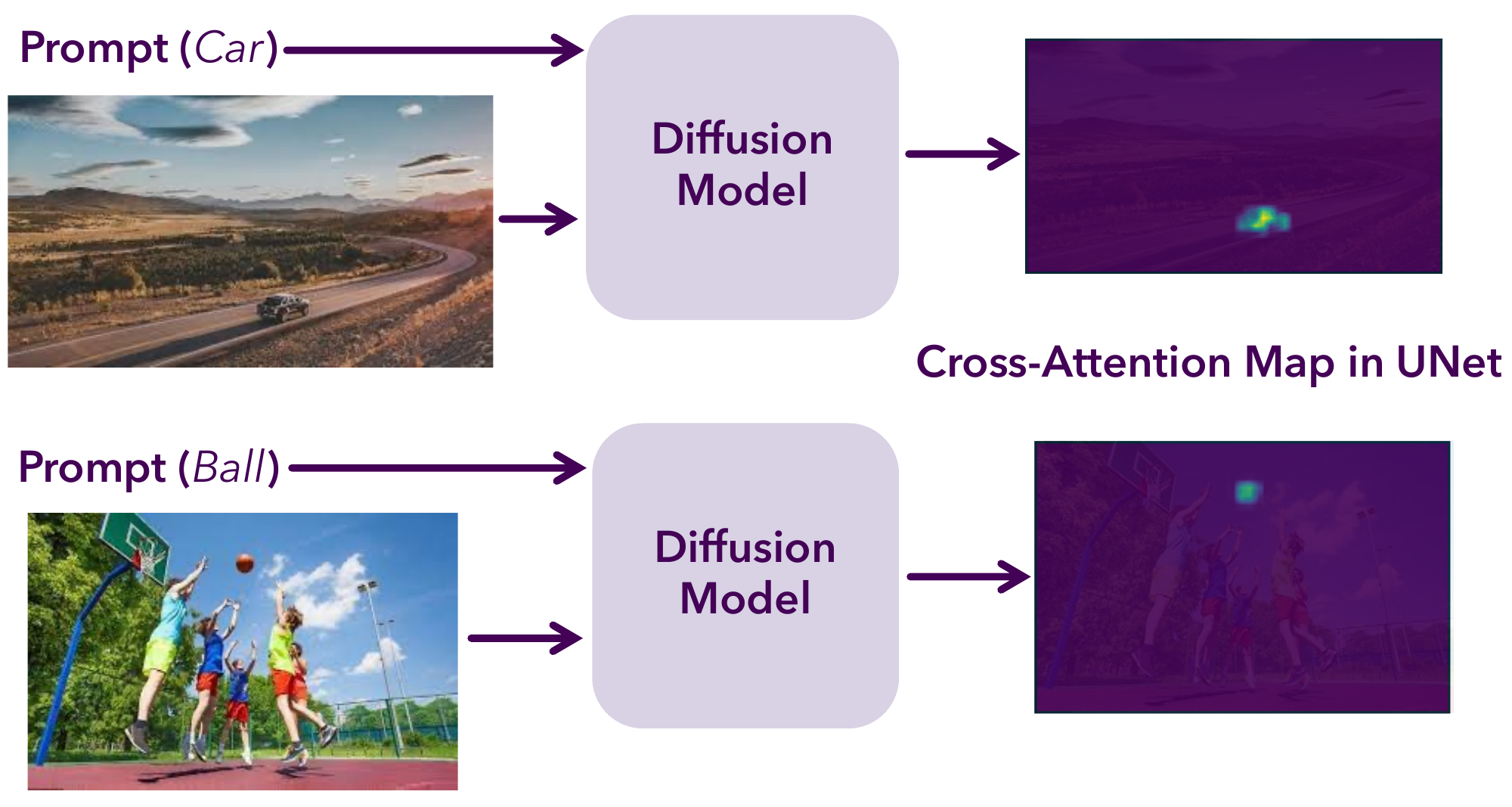}
\caption{To leverage the rich semantic knowledge embedded in the pre-trained text-to-image diffusion models~\cite{rombach2022high}, including their ability to capture semantic and structural information in video frames as well as contextual relationships across videos, we examine these models from a new perspective.
Specifically, we view the  diffusion models as a bridge connecting {their} input text prompts with the generated output images. This connection is explicitly manifested in the cross-attention maps within the UNet of the diffusion model, where text prompts can activate regions that share the same semantics. As shown in the figure, the model activates regions in the cross-attention maps that correspond to the semantics of input prompts (\textit{i.e.}, the ``\textit{Car}'' and ``\textit{Ball}'').}
\label{fig:vis_cross_attention}
\end{figure}

Meanwhile, pre-trained text-to-image diffusion models~\cite{khachatryan2023text2video,sun2024create,rombach2022high,peng2024harnessing} have achieved outstanding performance in various domains, including text-to-image generation~\cite{rombach2022high}, text-to-video generation~\cite{khachatryan2023text2video}, and cross-modal text-vision understanding~\cite{qin2025unicot}. For example, in the text-to-image domain, models such as Stable Diffusion~\cite{rombach2022high} have demonstrated remarkable capabilities in generating images that are diverse, rich in detail, and consistent with reasonable spatial structures, all guided by user-defined prompts. These results indicate that the pre-trained text-to-image diffusion models possess a comprehensive understanding of visual representations, spanning pixel-level semantic details and structural layouts, including object textures, shapes, and spatial arrangements. 
Moreover, recent studies~\cite{tang2023emergent} have shown that image diffusion models, even without training or fine-tuning on video data, exhibit strong capabilities in capturing contextual relationships within videos.
Given their capacity to understand the semantic and structural information of video frames and the contextual relationships within videos, a natural question arises:  can we leverage the rich knowledge embedded in the text-to-image diffusion models to perform unsupervised visual tracking?

Nevertheless, leveraging the knowledge implicitly embedded in the pre-trained text-to-image diffusion models for unsupervised object tracking in videos is non-trivial. These models are designed to generate images from text prompts and are not inherently suited for visual object tracking, which is a discriminative rather than a generative task. To address this challenge, we reinterpret the functionality of text-to-image diffusion models from a new perspective. Specifically, we view the diffusion model as a bridge linking the semantic information in input prompts with the visual content of output images. This semantic linkage is explicitly manifested in the model’s cross-attention maps~\cite{khani2023slime}, where text prompts activate regions (areas with high pixel values on the attention maps) that are semantically aligned with the prompts, as shown in~\cref{fig:vis_cross_attention}.
This observation suggests that, by learning a prompt that represents the tracking target defined by a given bounding box in the first frame, text-to-image diffusion models can be guided to highlight the target region in the corresponding cross-attention maps of the input video frames.
In this way, the rich knowledge embedded in pretrained text-to-image diffusion models can be leveraged to tackle the challenging problem of unsupervised visual object~tracking.

However, learning such a prompt that represents the target  is not straightforward for two main reasons: (i) In the diffusion model, cross-attention maps at different layers tend to capture different cues.  Specifically, intermediate layers are more inclined to represent semantic information, whereas deeper layers focus on texture details~\cite{zhang2023tale}. (ii)
In our unsupervised tracking setup, the only information provided about the target is its bounding box (appearance) in the first frame. However, as the target moves, both its appearance and its interaction with the background evolve over time~\cite{zhu2015weighted,fang2017spatial}. Therefore, maintaining continuous target tracking under these conditions constitutes our second challenge.

To address these challenges, we propose Diff-Tracking, which consists of two main components: an \textit{initial prompt learner} and an \textit{online prompt updater}.
The \textit{initial prompt learner} is designed to learn a prompt that represents the target through the cross-attention mechanism, enabling the diffusion model to perform object tracking. 
Besides, to ensure robust tracking even under significant appearance variations, we incorporate an attention harmonization method into the initial prompt learner. This method captures the target–background relationship as complementary cues to appearance, thereby enabling more reliable target localization.
We also design a fusion head that enables the initial prompt learner to effectively integrate information from heterogeneous cross-attention maps when learning the initial prompt, thereby exploiting their complementary characteristics.
Since the learned prompt is a target-specific embedding that represents only the tracking target in the current video, it may fail to capture traits shared across different targets, such as the observation that most tracked objects are salient~\cite{mahadevan2012connections,wang2024salient}. To improve the generalization of our method, we introduce a target-shared embedding that, together with the target-specific embedding, forms the final prompt produced by the initial prompt learner. It is worth noting that the target-shared embedding is learned jointly with the initial prompt learner during training, while the target-specific embedding is adapted for each video at test time using the bounding box of the target in the first frame.

The second component of our Diff-Tracking, \textit{i.e., online prompt updater,} is designed to update the learned initial prompt based on the target’s motion, thereby improving continuous tracking of the target. Specifically, in our online prompt updater, we employ a motion information extractor to capture motion of the target object by deriving target-conditioned motion information between the current frame and the previous one. However, relying solely on the motion information from two consecutive frames could be unreliable, as factors like occlusions and illumination changes may degrade the accuracy of such short-term motion cues and disrupt the spatio-temporal continuity of the captured motion~\cite{cheng2022implicit}. To address this issue, we further incorporate long-term motion information to enhance the spatio-temporal continuity of the motion representation in our offline tracking task. Moreover, when the target has a similar color to the background, such as tracking a black ant in a dark environment, it becomes difficult to distinguish the target from the background and to obtain reliable motion information. To address this, we convert the frames from the RGB domain to the frequency domain, extract motion information from the frequency-domain frames, and integrate it into the online prompt updater. This design leverages the observation that targets and backgrounds often exhibit distinct textures~\cite{ning2009robust,chase}, and frequency representations are especially effective in capturing such differences~\cite{hu2024sfdfusion}. Consequently, even when the target and background share similar colors, our model can more effectively distinguish them and extract more reliable motion information for the target.

We summarize our contributions:
\begin{itemize}
\item In this work, we approach the challenging unsupervised object tracking from a new perspective by leveraging the rich knowledge encoded in the pre-trained text-to-image diffusion models.
\item To adapt the text-to-image diffusion models, originally developed for image generation, to the tracking task, we introduce Diff-Tracking, a framework with two components: an initial prompt learner that generates a target-representative prompt, and an online prompt updater that refines this prompt using motion information to enable continuous tracking.
\item Our method achieves SOTA results on six unsupervised object tracking benchmarks, encompassing challenging scenarios such as severe occlusions (\textit{e.g.}, OTB 2015~\cite{7001050}) and long-term target tracking (\textit{e.g.}, LaSOT~\cite{fan2019lasot}). {Our contribution analysis further confirms that the gains stem from our tracking-specific designs rather than from large-scale pretraining alone.}
\end{itemize}

As an extension of our ECCV 2024 conference paper~\cite{zhang2024diff}, this work introduces several improvements in both methodology and experiments. \textit{\ding{172} Methodological improvements:}
(i) In the conference version of our method, the learned embedding that represents the target is essentially target-specific, encoding only the object in the current video. This design may fail to capture shared properties across targets, such as saliency~\cite{mahadevan2012connections,wang2024salient}. To address this issue, we introduce a target-shared embedding (see subsection Learning of the Initial Prompt in \cref{subsec:Initial Prompt Learner}), which is designed to capture common traits of tracking targets and thereby improve generalization.
(ii) The conference version relies on a single cross-attention map layer from the diffusion model’s UNet to learn the target embedding. However, different UNet layers capture complementary cues: intermediate layers focus on semantic structure, while deeper layers encode fine-grained textures~\cite{zhang2023tale}. To leverage this, we introduce an Attention Map Fusion head that integrates information from multiple layers (see subsection Attention Map Fusion Head in \cref{subsec:Initial Prompt Learner}).
(iii) The previous version often struggles when targets and backgrounds have similar RGB colors, resulting in unreliable motion extraction and less effective online prompt updates. To overcome this, we incorporate motion information from the frequency domain. Since targets and backgrounds often exhibit distinct textures~\cite{ning2009robust,chase}, frequency-domain representations are particularly effective at capturing such differences~\cite{hu2024sfdfusion}. This design enables our method to extract more robust motion cues in challenging scenarios, supporting more reliable online updates (see subsection RGB-to-Frequency Transformation in \cref{subsec:Online Prompt Updater}).
\textit{\ding{173} Experimental improvements:} We further conduct experiments on the challenging VOT2020 dataset~\cite{kristan2020eighth}, providing additional evidence of the effectiveness of our method (see \cref{subsec:main results}).
In addition, we extend comparisons against a wider range of state-of-the-art trackers (see \cref{subsec:main results}). Finally, we conduct new ablation studies and provide deeper insights into the effectiveness of our proposed modules, as demonstrated by the experimental results (see \cref{subsec:ablation}).

The rest of this paper is organized as follows. Section \ref{sec:related} first reviews the current progress in supervised and unsupervised visual object tracking, as well as advances in text-to-image diffusion models. Next, section \ref{sec:Preliminaries} introduces the preliminaries of text-to-image diffusion models, providing the foundation for our approach. Section \ref{sec:method} then presents the detailed design of our method together with the motivations behind it. Following this, section \ref{sec:experiments} reports the main experimental results along with ablation studies. Finally, section \ref{sub:conclusion} concludes the paper.

\section{Related Work}
\label{sec:related}

\subsection{Supervised Visual Object Tracking}
Supervised visual object tracking~\cite{shen2021distilled,li2015nus} is a core problem in computer vision, with wide-ranging applications in areas such as autonomous driving~\cite{chen2021novel,gao2019manifold}, human behavior analysis~\cite{li2021uavhuman}, and robotics~\cite{papanikolopoulos1993visual,budiharto2020design}.
In recent years, deep learning-based trackers have achieved remarkable performance gains. These methods can be broadly categorized into two groups: online-optimized trackers~\cite{danelljan2017eco,danelljan2019atom,DiMP,danelljan2020probabilistic} and offline trackers~\cite{li2019siamrpn++,yu2020deformable}.
Among them, online-optimized trackers rely on carefully designed online update mechanisms~\cite{hu2023siammask}. Specifically, they first estimate the rough position of the target using ridge regression with updated template kernels, followed by refinement to produce accurate bounding boxes.
In contrast, offline trackers learn to match the template and the search region in a metric space, and this category is dominated by Siamese-network based approaches~\cite{bertinetto2016fully,li2019siamrpn++}. The pioneering work SiamFC~\cite{bertinetto2016fully} extracts features of template and search patches with a shared backbone network and applies cross-correlation to generate a response map for target localization. Besides, a variety of efforts have been made to enhance Siamese-based trackers, including stronger backbone networks~\cite{li2019siamrpn++} and effective template-search fusion strategies~\cite{l2fuse}. Efficient tracking has also been explored through  network architecture search~\cite{EffSearch}, and lightweight designs~\cite{lightrack}. Despite these advances, all of the above methods rely on extensive supervised training with large-scale annotated video datasets to learn the correspondence between template and search regions. In contrast, our work presents a new unsupervised visual object tracking framework that leverages the rich knowledge embedded in the pre-trained diffusion model, thus eliminating the need for costly annotations.

\subsection{Unsupervised Visual Object Tracking}
Since collecting video annotations is costly~\cite{javed2022visual}, unsupervised visual object tracking has drawn considerable attention from the research community~\cite{wang2019unsupervised,sio2020s2siamfc, zheng2021learning}. The pioneering work UDT~\cite{wang2019unsupervised} develops an unsupervised tracker based on Discriminative Correlation Filters~\cite{zuo2018learning}, which is trained using forward-backward tracking of frames guided by a consistency loss.
 Besides, Sio \textit{et al.}~\cite{sio2020s2siamfc} introduce a Siamese network based unsupervised training framework that extracts self-supervision from single frames, where adversarial masking is learned to construct template–search pairs within the same frame. However, the performance of these methods~\cite{sio2020s2siamfc, wang2019unsupervised} {relies} heavily on online updating schemes. Without online updates, these unsupervised trackers struggle to handle challenging objects with large variations.
Recently, leveraging self-supervision signals from both spatial and temporal dimensions has emerged as a promising direction for unsupervised tracking. Specifically, Zheng \textit{et al.}~\cite{zheng2021learning} propose a two-stage approach that first performs naive training on single-frame pairs, followed by cycle training to capture longer temporal dependencies. Wu \textit{et al.}~\cite{pul} introduce a method that initially learns a background discrimination model through contrastive learning, and then continues training with temporally corresponding patches using a noise-robust loss. In addition, several pretext tasks~\cite{timecycle,videocp,foo2023system} have been designed to learn visual representations from videos by exploiting the idea of forward-backward tracking. Following~\cite{sio2020s2siamfc}, state-of-the-art unsupervised trackers~\cite{zheng2021learning,shen2022unsupervised} also employ similar siamese network structure. In contrast, this paper performs unsupervised visual object tracking from a novel perspective by leveraging the rich prior knowledge embedded in a text-to-image diffusion model pre-trained on large-scale data.

\subsection{Text-to-Image Diffusion Models}
Diffusion models~\cite{saharia2022photorealistic,chen2023pixart,foo2024action,zhou2026frequency} have achieved remarkable progress in recent years, largely due to their ability to generate highly realistic visuals when trained on large-scale datasets.
{Representative examples include text-to-image diffusion models}, which demonstrate strong capabilities in generating diverse images from text prompts~\cite{rombach2022high}. Building on their strong performance, a growing body of research has leveraged the rich prior knowledge embedded in these models to advance tasks such as image editing~\cite{kawar2023imagic}, text-to-video generation~\cite{khachatryan2023text2video}, and video editing~\cite{zhang2025visual}. These areas have been reshaped by the versatility and effectiveness of the works built on the pre-trained diffusion models. {Beyond generation, a growing line of research has explored the attention mechanisms of diffusion models for understanding and manipulating visual content. Hertz et al.~\cite{hertz2022prompt} show that cross-attention maps in diffusion models control the spatial layout of generated images and enable fine-grained image editing by directly manipulating these maps. Chefer et al.~\cite{chefer2023attend} further guide cross-attention activations during the generation process to improve alignment between generated images and text prompts. Tang et al.~\cite{tang2023daam} provide a systematic analysis demonstrating that cross-attention maps serve as reliable attribution maps that link text tokens to corresponding image regions. In addition, Tumanyan et al.~\cite{tumanyan2023plug} leverage the self-attention features of diffusion models for image-to-image translation, revealing that these features encode rich spatial and semantic structure. These works collectively establish that the attention maps of diffusion models are well-grounded signals for spatial localization and semantic correspondence, which provides a strong foundation for our use of cross-attention maps as target localization signals in tracking.} In contrast to these efforts, we leverage rich prior of the pre-trained text-to-image diffusion models to tackle the challenging problem of unsupervised visual object~tracking.

\section{Background: Text-to-Image Diffusion Models}
\label{sec:Preliminaries}

Unsupervised visual object tracking is a challenging task that requires a strong understanding of the semantics and structures of video frames, as well as the contextual relationships within videos~\cite{wang2021unsupervised,sio2020s2siamfc,zheng2021learning}. Since pre-trained text-to-image diffusion models can generate content aligned with the semantics of user-provided prompts, they exhibit a powerful understanding to capture image content and structure~\cite{rombach2022high}. Furthermore, even without additional training, they demonstrate a strong capacity to model contextual relationships within videos~\cite{tang2023emergent}. In this paper, we aim to leverage the capabilities of pre-trained text-to-image diffusion models for the unsupervised visual object tracking. Specifically, we exploit the cross-attention mechanism of the diffusion model by learning a prompt that represents the tracking target, which in turn highlights the regions corresponding to the target on the cross-attention maps. Moreover, as the target moves, its appearance may vary, making it crucial to exploit the semantic relationship between the target and background as auxiliary information for tracking. To incorporate this relationship and enhance robustness of our method, we leverage the self-attention mechanism of the diffusion model, as its self-attention maps encode semantic relationships among pixels.

In the following, we first present a detailed introduction to the training process of text-to-image diffusion models, followed by descriptions of their cross-attention and self-attention mechanisms. For clarity, we use the widely adopted Stable Diffusion model~\cite{rombach2022high} as an illustrative example.

\subsection{Training Process}
Text-to-image diffusion models are designed to reconstruct an image from random Gaussian noise through a denoising process over multiple timesteps. This is accomplished by training a noise prediction model to progressively remove the noise via a multi-step reverse diffusion process conditioned on a text prompt. Specifically, during the training process, noise is added to the latent representation of the input image at each timestep, and the noise prediction model is tasked with predicting the noise introduced at the current timestep from the encoded image. The training objective is formulated as:
\begin{equation} \label{eq:diffusion model} L_\mathrm{DM} = \mathbb{E}_{\mathcal{E}_\mathrm{img}(I), \epsilon \sim \mathcal{N}(0,1), t}\left[\| \epsilon - \epsilon_{\theta}\left(z_t, t, \mathcal{E}_\mathrm{txt}(p) \right)\|_2^2\right]. \end{equation}

Here, $I$ is the input image encoded by an image encoder $\mathcal{E}_\mathrm{img}$.  $\epsilon$ is the added noise, sampled from a standard Gaussian distribution $\mathcal{N}(0,1)$. $\epsilon_{\theta}$ represents the noise prediction model parameterized by $\theta$ (typically implemented as a UNet~\cite{ronneberger2015u}), which is trained to estimate the added noise at timestep $t$, conditioned on the text prompt $p$ and the image latent $z_t$.
 The text prompt $p$ is encoded by the text encoder $\mathcal{E}_\mathrm{txt}$.

\subsection{Cross-attention  Mechanism}
In the diffusion models, the interaction between image content and the text prompt is captured through the cross-attention layers of the denoising UNet. Specifically, in each cross-attention layer, the query $Q_c$ is computed from the noisy image latent, while the key $K_c$ and value $V_c$ are derived from the text embedding. Cross-attention map $M_c$ is given by:
\begin{equation}
\label{eq:m_c}
M_c = \operatorname{Softmax}\left(\frac{Q_c K_c^T}{\sqrt{d}}\right),
\end{equation}
where $d$ denotes the projection dimension of $Q_c, K_c$, and $V_c$. By correlating the visual features in $Q_c$ with the semantic features in $K_c$, the cross-attention map $M_c$ identifies image regions that align with the text prompt. This property allows us to leverage diffusion models for visual object tracking by learning a target prompt via the cross-attention mechanism and using it to activate the corresponding region in the cross-attention map of the input video frame.

\subsection{Self-Attention Mechanism}
In the architecture of the diffusion model, alongside the cross-attention layers, self-attention layers are also integrated within the UNet. These layers play a key role in capturing semantic correlations among pixels in an image. Specifically, the self-attention map $M_s$ is defined as:
\begin{equation}
\label{eq:m_s}
M_s = \operatorname{Softmax}\left(\frac{Q_s K_s^{T}}{\sqrt{d}}\right),
\end{equation}which shares the same form as the cross-attention map $M_c$. 
The difference is that the query $Q_s$, key $K_s$, and value $V_s$ in $M_s$ are all derived from the noisy image latent representation. Because the computation of the self-attention map $M_s$ involves correlations between $Q_s$ and $K_s$, both obtained from image pixels, $M_s$ captures the semantic relationships among pixels.

\begin{figure*}[!t]
\centering
\includegraphics[width=6.8in]{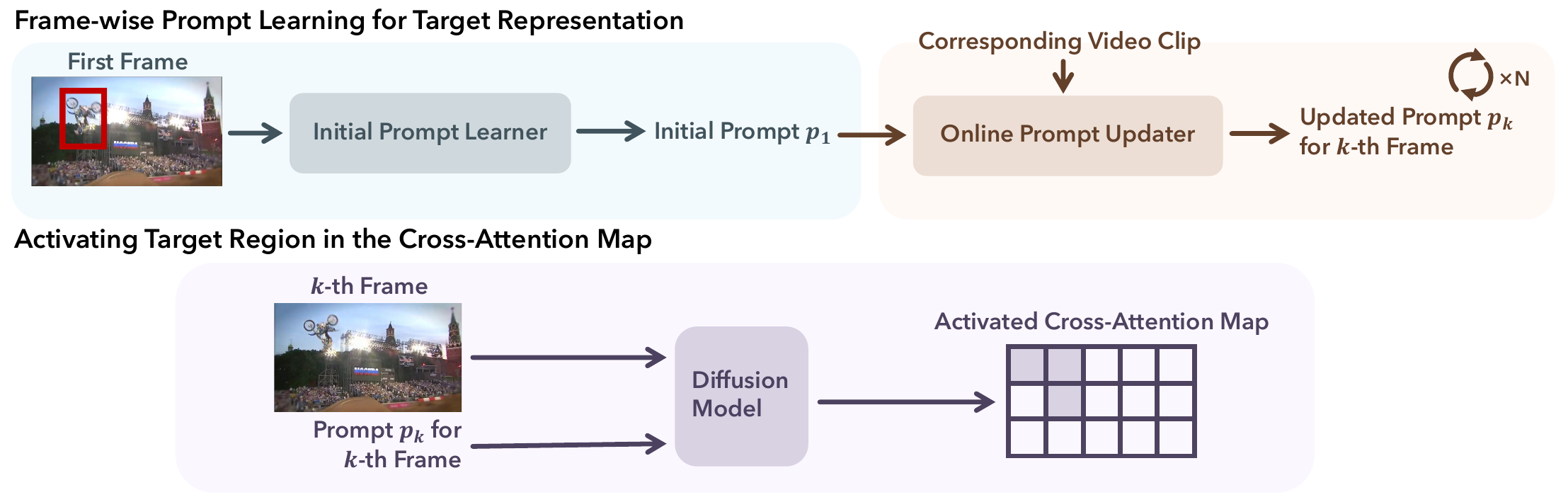}
\caption{The overall architecture of our Diff-Tracking framework. To harness the rich knowledge embedded in pre-trained diffusion models for the challenging task of unsupervised visual tracking, we propose learning a target prompt through the model’s cross-attention mechanism and then using the learned prompt to activate the target region in the cross-attention map of each input frame. Our framework comprises two main components: an initial prompt learner and an online prompt updater. Specifically, the initial prompt learner generates an initial prompt from the provided bounding box of the target in the first frame using.The online prompt updater then refines this prompt using target motion extracted from the corresponding video clip to ensure continuous tracking across frames. For the inference of our method at frame $k$, we feed both the $k$-th frame image and its corresponding prompt $p_k$ to the diffusion model, which activates the target region through the resulting cross-attention map. The initial prompt tracks the target for frames~1--5. Starting from frame~6, the online prompt updater refines the prompt at every frame through the end of the video; we denote this process as $\times N$ iterations in the figure, where $N = L - 5$ for a video of $L$ frames.}
\label{fig:pipeline}
\end{figure*}

\section{Method}
\label{sec:method}
Unsupervised visual object tracking is a challenging task that requires the tracker to follow an arbitrary target without training on video data with ground-truth annotations. Specifically, the target is specified by a bounding box in the first frame, and the tracker is required to accurately localize it in all subsequent frames of the sequence based on this bounding box. In this work, we address the challenging task from a novel perspective by leveraging the rich knowledge contained in pretrained text-to-image diffusion models for unsupervised visual tracking. Yet, such diffusion models are originally designed to synthesize images from text prompts, which makes them appear unsuitable for performing tracking. To handle it, we reinterpret the diffusion model as a bridge between the text and image modalities. This connection is realized through the cross-attention mechanism in the UNet of the diffusion model.

Specifically, when both a prompt (text) and an image are simultaneously input into the diffusion model, the input prompt can activate semantically corresponding regions in the cross-attention map of the input image on the diffusion model's UNet (see \cref{fig:vis_cross_attention}). Hence, building on this insight, we introduce Diff-Tracking, a framework that leverages the cross-attention mechanism to utilize rich knowledge embedded in the pre-trained diffusion models for the challenging unsupervised object tracking. As illustrated in \cref{fig:pipeline}, our Diff-Tracking framework is composed of two core components: an initial prompt learner and an online prompt updater. The initial prompt learner is designed to acquire a prompt that can activate the target's bounding box in the first frame via the cross-attention mechanism. Then, the updater dynamically adapts this learned prompt based on the target's motion, thereby enabling continuous tracking throughout the sequence.

In the subsequent sections, we first elaborate on our proposed initial prompt learner (\cref{subsec:Initial Prompt Learner}), followed by a detailed description of the online prompt updater (\cref{subsec:Online Prompt Updater}). We conclude by outlining the complete training and testing pipelines of our approach (\cref{subsec:training_and_testing}).

\subsection{Initial Prompt Learner}
\label{subsec:Initial Prompt Learner}

\begin{figure*}[!t]
\centering
\includegraphics[width=7.1in]{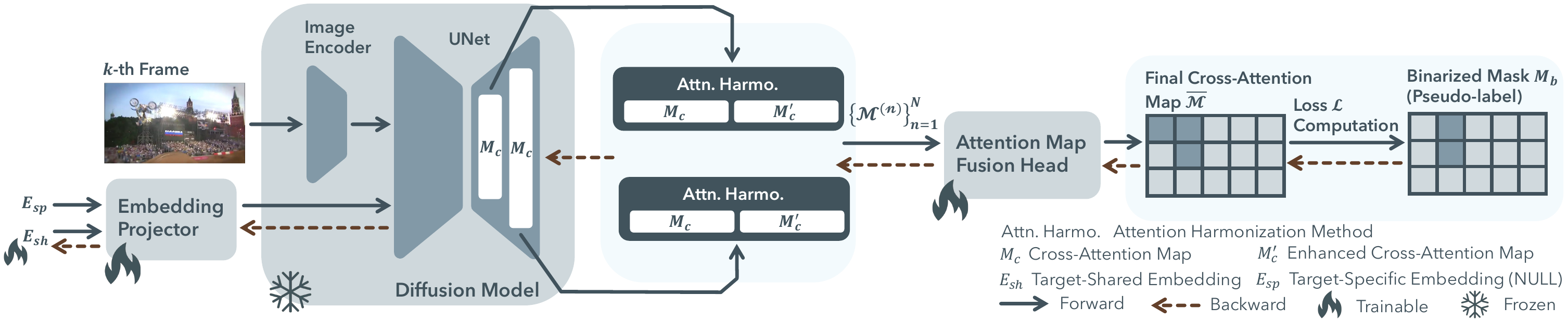}
\caption{The learning process for the target-shared embedding $E_{sh}$ in our initial prompt learner. Specifically, given the $k$-th frame and its corresponding binarized mask $M_b$ generated as pseudo-labels following USOT~\cite{zheng2021learning}, we concatenate the target-shared embedding $E_{sh}$ with the target-specific embedding $E_{sp}$ (set to NULL during this phase) and project them through an embedding projector to match the dimensionality of the diffusion model. The projected embedding, along with the $k$-th frame, is then fed into the diffusion model, from which cross-attention maps $M_c$ activated by the embedding are extracted across different decoder layers of the UNet. These maps $M_c$ are enhanced using self-attention maps that capture pixel-wise relationships, yielding enhanced cross-attention maps $M_c^\prime$ (see \cref{eq:enhanced attention map}). Through our attention harmonization method, we then fuse $M_c$ and $M_c^\prime$ to produce the output attention maps $\mathcal{M}$ (\cref{eq:output cross-attention map}). These attention maps $\left\{\mathcal{M}^{\left(n\right)}\right\}_{n=1}^N$ of different $N$ layers are then aggregated via the attention map fusion head to generate the final attention map $\overline{\mathcal{M}}$ (\cref{eq:calculation of the final attention map}). We compute the loss $\mathcal{L}$ between $\overline{\mathcal{M}}$ and the pseudo-label $M_b$ (see \cref{eq:loss of attention map}), then update the attention map fusion head, embedding projector, and $E_{sh}$ via backpropagation while keeping the diffusion model frozen. 
For learning $E_{sp}$, we employ the same architecture but update only $E_{sp}$, using the target's bounding box from the first frame instead of the binarized mask. During inference, our method reuses this initial prompt learner architecture.}
\label{fig:initial_prompt_learner}
\end{figure*}

The initial prompt learner enables the text-to-image diffusion model to identify the tracking target by learning {a prompt that represents the target},  and the prompt can activate the corresponding region of the target in the diffusion model's cross-attention map. Specifically, our initial prompt learner consists of two core components: an attention harmonization method and an attention map fusion head (see \cref{fig:initial_prompt_learner}). The attention harmonization head integrates target-background relationships into the learned initial prompt to address common tracking challenges including severe deformation, occlusion, and visually similar distractors~\cite{zhu2015weighted}, as these spatial relationships provide critical cues for robust tracking under challenging conditions~\cite{fang2017spatial}. Additionally, since cross-attention maps across UNet depths capture complementary information~\cite{zhang2023tale}, with intermediate layers encoding semantic structure and deeper layers capturing fine-grained textures, we design an attention map fusion head to aggregate the cross-attention maps, leveraging the heterogeneous signals to enhance tracking performance.

With these components, our initial prompt learner can generate the initial prompt. However, this prompt remains a target-specific embedding that encodes only the current target's characteristics, potentially overlooking common properties shared across tracking targets, such as visual saliency~\cite{mahadevan2012connections,wang2024salient}. To enhance robustness and generalization, we introduce a target-shared embedding that combines with the learned target-specific embedding to form the final initial prompt.

Below, we first present the attention harmonization method, then describe the attention map fusion head, and finally detail the learning of the target-specific embedding and the target-shared embedding that together constitute our initial prompt.

\noindent \textbf{Attention Harmonization Method.}  To encode the relationship between the target and its background within the learned prompt, we develop an attention harmonization method  that enhances the cross-attention map with the self-attention map, since the self-attention map in the pre-trained text-to-image diffusion model captures semantic relationships among pixels (as discussed in Section~\ref{sec:Preliminaries}).

Specifically, for the cross-attention map $M_c$ extracted from the UNet (see Eq. \ref{eq:m_c}), each pixel in $M_c$ can be associated with a self-attention map $M_s$ (see Eq. \ref{eq:m_s}) to reflect its relationship with other pixels. To encode such inter-pixel relationships, we enhance $M_c$ using $M_s$. The enhanced cross-attention map $M_c^\prime$ is computed as:
\begin{equation}
\label{eq:enhanced attention map}
M_c^\prime(:,:)=\sum_{i=1} \sum_{j=1} M_c(i, j) \cdot M_s(i, j,:,:),
\end{equation}
where $M_c(i, j)$ denotes the attention value at position $(i, j)$ on $M_c$, and $M_s(i, j,:,:)$ represents the self-attention map associated with pixel $(i, j)$ of $M_c$.

After obtaining $M_c^\prime$, which encodes inter-pixel relationships, we integrate it with the original cross-attention map $M_c$ to form the final output cross-attention map $\mathcal{M}$. To do so, we resize $M_c^\prime$ to match the size of $M_c$ and perform an element-wise weighted summation of the two maps. The output cross-attention map $\mathcal{M}$ is defined as:
\begin{equation}
\label{eq:output cross-attention map}
\mathcal{M} = (1 - \alpha) \cdot M_c^\prime + \alpha \cdot M_c,
\end{equation}
where $\alpha$ is a learnable parameter that balances the contributions of $M_c^\prime$ and $M_c$.

\noindent  \textbf{Attention Map Fusion Head.}
To leverage cross-attention maps across layers that capture complementary properties (for example, intermediate layers emphasize semantic structure while deeper layers capture fine-grained textures~\cite{zhang2023tale}), we propose an attention map fusion head $H_{\mathrm{f}}$ that enables the initial prompt to integrate information from multiple layers during learning. Given the symmetry between the encoder and decoder in the UNet of diffusion models, our method leverages the attention maps from the decoder. Specifically, we upsample the output cross-attention attention maps $\left\{\mathcal{M}^{\left(n\right)}\right\}_{n=1}^N$ from $N$ layers to a uniform spatial size, feed them into the  $H_{\mathrm{f}}$, and compute a weighted combination of the outputs to obtain the final attention map $\overline{\mathcal{M}}$ for learning the initial prompt. The entire process can be formulated as follows:
\begin{equation}
\label{eq:calculation of the final attention map}
\overline{\mathcal{M}}= \frac{1}{N} 
 \sum_{n=1}^N w^{\left(n\right)} \cdot H_{\mathrm{f}}\left(\mathcal{M}^{\left(n\right)}\right).
\end{equation}
In this formulation, $w^{(n)}$ denotes the learnable weight assigned to the $n$-th attention map produced by $H_{\mathrm{f}}$.

\noindent \textbf{Learning of the Initial Prompt.} After obtaining the final attention map from the fusion head, we employ this map to guide the learning of the initial prompt. It is worth noting that for computational efficiency, we represent the learned prompt as an embedding vector rather than a textual sentence. Specifically, our initial prompt is composed of a \textit{target-specific embedding} and a \textit{target-shared embedding}.  The \textit{target-specific embedding} is designed to represent the target to be tracked in the current video. During inference, it is adapted for each video through a test-time adaptation procedure, where it is updated using the bounding box annotation from the first frame. Besides, the \textit{target-shared embedding} captures properties common to diverse tracking targets, such as the fact that most tracked objects are salient in videos. This embedding enhances the robustness of the approach and, unlike the target-specific embedding, remains frozen during inference after training.
When both target-shared and target-specific embeddings are available, we concatenate them and project them to the diffusion model's input dimensions using an embedding projector, which is implemented as a lightweight two-layer MLP. These projected embeddings are then fed into the diffusion model, where the cross-attention mechanism identifies the tracking target region.
Below, we describe the learning process of the target-shared embedding and the target-specific embedding in detail.

\noindent \textit{\ding{172} Target-Shared Embedding $E_{sh}$.} To learn the target-shared embedding, we follow the previous unsupervised visual tracking method USOT~\cite{zheng2021learning} to generate pseudo-labeled training data. Specifically, similar to USOT, we exploit the observation that the motion patterns of foreground objects and background regions in a video often differ. Hence, we use the unsupervised optical flow ARFlow~\cite{liu2020learning} to locate regions (boxes) corresponding to foreground objects, and then apply dynamic programming to select temporally consistent box sequences from these candidates. After obtaining the box sequences, \textit{i.e.}, {the generated pseudo-labels}, we use them to learn the target-shared embedding. The learning process consists of three steps. \textit{First}, following the workflow of the diffusion model, we feed the images together with a randomly initialized target-shared embedding into the model. \textit{Second}, we extract the cross-attention maps associated with the images and the target-shared embedding, and apply the attention harmonization mechanism and the fusion head to produce the final attention map. \textit{Third}, we compute the loss between the activated regions of the extracted attention map and pseudo labels of the corresponding images, and update $E_{sh}$.

Specifically, \textit{i)} to feed the frames with pseudo bounding boxes into the diffusion model, we follow  workflow of the diffusion model and first use the image encoder ($\mathcal{E}_{img}$ in \cref{eq:diffusion model}) to encode these frames into the latent space. We then add noise to the encoded representation to obtain the noisy latent representation. This representation is passed into the denoising UNet together with the target-shared embedding and the target-specific embedding, where the target-shared embedding is randomly initialized and the target-specific embedding is set to NULL.
 \textit{ii)} Then, to obtain the final attention map,  ($\overline{\mathcal{M}}$ in \cref{eq:calculation of the final attention map}), we first extract the cross-attention map from the UNet and enhance it using the self-attention map. The original cross-attention map and the enhanced one (\cref{eq:enhanced attention map}) are fused through our attention harmonization method to form the output cross-attention map ($\mathcal{M}$ in \cref{eq:output cross-attention map}). We then feed the cross-attention map into the fusion head to produce the final attention map (\cref{eq:calculation of the final attention map}). \textit{iii)} Finally, to learn the target-shared embedding, we compute the loss between the activated regions of the final attention map and the obtained pseudo labels (bounding boxes), and update the target-shared embedding through backpropagation. The loss is calculated as:
\begin{equation}
\label{eq:loss of attention map}
\mathcal{L}=\frac{1}{H W} \sum_{i=1}^H \sum_{j=1}^W\left(M_b(i, j)-\overline{\mathcal{M}}(i, j)\right)^2+ L_\mathrm{DM}.  
\end{equation}

In this equation, $M_b$ denotes a binarized mask with the same size as the input frame that contains our generated bounding box. Pixels inside the box are set to 1 and those outside are set to 0. $H$ and $W$ denote the mask height and width, respectively. $M_b(i,j)$ and $\overline{\mathcal{M}}(i,j)$ denote the pixel values at location $(i,j)$ in the mask $M_b$ and in the final attention map $\overline{\mathcal{M}}$, respectively.
Besides, $L_{DM}$ denotes the diffusion model loss (see \cref{eq:diffusion model}), which ensures that the learned target-shared embedding {$E_{sh}$} remains within the text embedding space interpretable by the diffusion model. Note that during the learning of the target-shared embedding, the target-specific embedding is never updated and remains empty. The parameters of the diffusion model are also kept fixed, while the parameters of both the embedding projector and the fusion head are updated.

\noindent \textit{\ding{173} Target-Specific Embedding $E_{sp}$.} After learning the target-shared embedding $E_{sh}$, we proceed to describe the learning of the target-specific embedding $E_{sp}$. In our approach, $E_{sp}$ is updated for each video so that, when concatenated with $E_{sh}$, the resulting initial prompt can accurately activate the target’s bounding box through the cross-attention mechanism. Specifically, $E_{sp}$ is updated only at inference time. Its learning follows the same procedure as $E_{sh}$: \textit{i)} we take the first frame, the learned $E_{sh}$, and a randomly initialized $E_{sp}$, pass them through the embedding projector to form the initial prompt, and feed it into the diffusion model. \textit{ii)}  We then obtain the output cross-attention map  $\overline{\mathcal{M}}$ in \cref{eq:loss of attention map}) using the proposed attention harmonization and the fusion head. \textit{iii)} We follow \cref{eq:loss of attention map} to compute the loss and backpropagate it to update $E_{sp}$. Notably, during inference-time learning of $E_{sp}$, only $E_{sp}$ is updated, while all other components, including $E_{sh}$ and the embedding projector, remain frozen.

\begin{figure}[!t]
\centering
\includegraphics[width=3.5in]{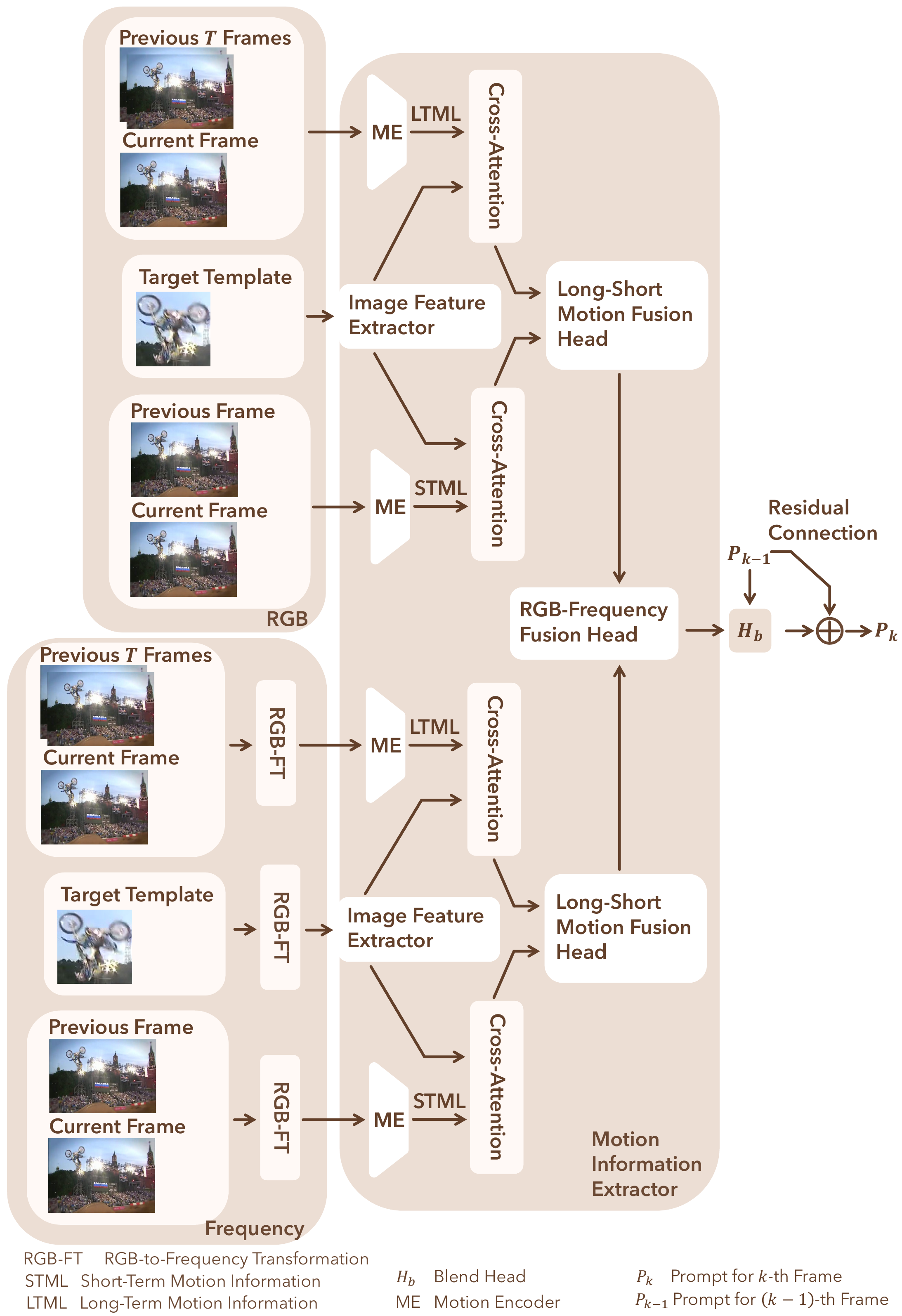}
\caption{The framework of our online prompt updater. The online prompt updater updates prompts based on target motion, enabling continuous tracking in our Diff-Tracking framework. Specifically, in the updater, we design a motion information extractor that processes both RGB and frequency modalities. The frequency modality enhances tracking in challenging scenarios where targets and backgrounds share similar colors but differ in texture~\cite{ning2009robust,chase}, as frequency representations effectively capture texture differences~\cite{hu2024sfdfusion}. Within each modality, we extract both short-term motion information reflecting real-time appearance changes and more robust long-term motion information. We employ cross-attention method to ensure the extracted motion features focus on the target. A long-short motion fusion head combines temporal motion cues, while a RGB-frequency fusion head integrates cross-modal information. The fused motion information is processed through a blend head to update the previous frame's prompt $p_{k-1}$ to generate the current frame's prompt $p_k$. We incorporate a residual connection to enhance the robustness of our updater.}
\label{fig:online_prompt_updater}
\end{figure}

After learning the target-specific embedding $E_{sp}$, we concatenate it with the target-shared embedding $E_{sh}$ and pass them through the embedding projector to obtain the initial prompt, which activates the bounding box of the tracking target on the first frame via the cross-attention mechanism.

\subsection{Online Prompt Updater}
\label{subsec:Online Prompt Updater}
After obtaining the initial prompt from the first frame, we can perform visual object tracking through the diffusion model's cross-attention mechanism. While ideally this prompt would suffice for tracking throughout the entire sequence, motion-induced appearance variations often lead to tracking failures when the same prompt is applied across frames. To address this issue, we propose an online prompt updater (see \cref{fig:online_prompt_updater}) that uses the target’s motion information to dynamically update the prompt for subsequent frames.

\begin{figure}[!t]
\centering
\includegraphics[width=3.5in]{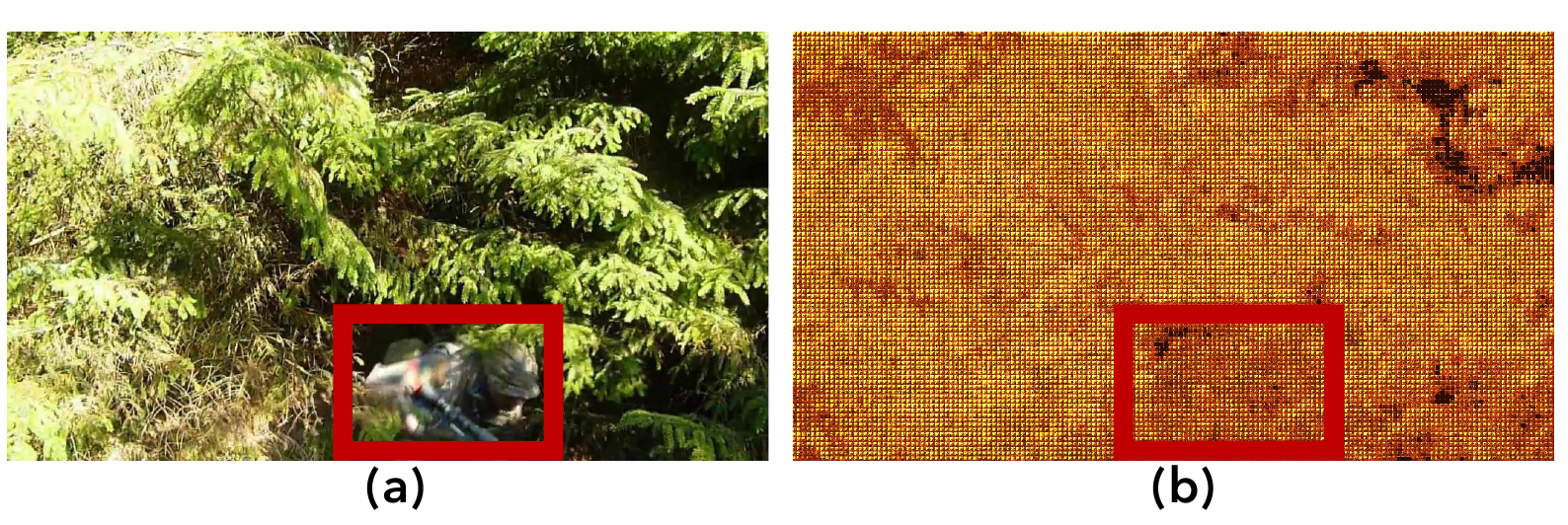}
\caption{Figures (a) and (b) show the visualizations of the RGB domain and the frequency domain (Y component), respectively, for a scene in the VOT 2020 dataset~\cite{kristan2020eighth} where a soldier in camouflage is being tracked. As observed, in the RGB domain the soldier is difficult to distinguish from the background due to the camouflage. In contrast, the frequency domain enables clearer separation of the soldier from the background, since the textures of the soldier and the background (trees) differ and frequency representations effectively capture these differences~\cite{hu2024sfdfusion}.}
\label{fig:frequency}
\end{figure}

In this updater, we exploit both short-term motion information capturing real-time object dynamics and more robust long-term motion information.
Specifically, to capture such real-time motion, we use the information between the current frame and the previous frame as the basis for updating the prompt. However, due to occlusions and illumination changes, such short-term motion derived from two consecutive frames often lacks strong spatio-temporal coherence. To address this limitation, we further incorporate long-term motion information into the update process, motivated by the observation that the long-term motion of moving objects generally exhibits strong spatio-temporal continuity~\cite{cheng2022implicit}.
Furthermore, to achieve accurate motion estimation in challenging scenarios where the target and background share similar colors and are thus difficult to distinguish, we incorporate frequency-domain motion information into the online prompt updater. This design leverages the observation that targets and backgrounds often differ in texture~\cite{ning2009robust,chase}, and frequency representations are particularly effective at capturing such differences~\cite{hu2024sfdfusion}. Consequently, even when the target and background appear visually similar, they can still be separated in the frequency domain (see \cref{fig:frequency}), enabling the extraction of reliable motion information for the target.

Generally, in our online prompt updater,  we first transform the input video frames from the RGB domain to the frequency domain to extract motion information of the tracking target in the frequency domain. We then apply the motion information extractor to separately capture long-term and short-term motion of the target from input frames in both the RGB and frequency domains. These motion cues update the prompt and enable continuous tracking through the diffusion model’s cross-attention mechanism.

We first introduce the RGB-to-frequency transformation, then present the design of the motion information extractor, and finally describe the process of online prompt updating.

\noindent \textbf{RGB-to-Frequency Transformation.} 
To compute motion information in the frequency domain, we first convert the frames from the RGB domain to the frequency domain based on Fourier analysis~\cite{wen2022high}. According to Fourier analysis, any finite-energy signal, such as an image, can be expressed as a linear combination of sine and cosine functions (\textit{i.e.}, frequency bases), and this transformation is typically implemented using the Discrete Cosine Transform (DCT)~\cite{wen2022high}. Hence, following Fourier analysis, we apply DCT to carry out the RGB-to-frequency transformation.

Specifically, for a frame of size $m \times n$, its conversion from the RGB domain to the frequency domain using DCT is defined as follows:
\begin{equation}
\begin{aligned}
Y(u, v) &=  \frac{2}{\sqrt{M \times N}} C(u) C(v) \\ & \sum_{m=0}^{M-1} \sum_{n=0}^{N-1} X(m, n) \cos \frac{(2 m+1) u \pi}{2 M} \cos \frac{(2 n+1) v \pi}{2 N},
\end{aligned}
\end{equation}
where $X(m,n)$ represents the pixel value of the input frame in the spatial domain, and $Y(u,v)$ represents the DCT coefficient of the output frame in the frequency domain. Besides, $u=1,2, \ldots, M-1 ; v=1,2, \ldots, N-1$, and
\begin{equation}
C(u) = 
\begin{cases}
\frac{1}{\sqrt{2}}, & u=0 \\
1, & u \neq 0
\end{cases}
\;
,
\quad
C(v) = 
\begin{cases}
\frac{1}{\sqrt{2}}, & v=0 \\
1, & v \neq 0
\end{cases}
\;
.
\end{equation}

\noindent \textbf{Motion Information Extractor.} Following the RGB-to-frequency domain transformation, we extract motion information from both RGB and frequency domains to update the learned prompt, enabling continuous target tracking. The extracted motion information includes short-term cues that capture the target’s instantaneous movement and long-term cues that reflect its movement over an extended period. Below, we first describe the extraction of motion information in the RGB domain, and then explain how motion information is extracted in the frequency domain.

Specifically, to extract motion information in the RGB domain, we input the current frame and the previous frame into a motion encoder to obtain short-term motion information. In addition, we stack the current frame with the $T$ preceding frames as a video sequence and feed them into another motion encoder to obtain long-term motion information.  However, these two types of features are extracted from the entire image, while our goal is to obtain target-conditioned motion. To address this, we employ cross-attention method to establish connections between the target and extracted motion features, generating target-conditioned motion representations.

In particular, the first cross-attention method is constructed using the target’s appearance features as the query. These features are obtained by feeding the target template from the first frame into the image feature extractor, while the long-term motion information serves as the key and value. Similarly, another cross-attention method is built using the same target appearance features as the query, with the short-term motion information as the key and value. To integrate these motion features, we design a long-short motion fusion head, implemented as a lightweight multi-layer perceptron (MLP) consisting of two fully connected layers and a ReLU activation. This fusion head takes both the long-term and short-term motion information as input and produces the fused motion representation.

It is worth noting that the extraction of motion information in the frequency domain follows the same procedure as in the RGB domain, with the only difference being that the inputs are represented in the frequency domain rather than the RGB domain. To avoid redundancy, we do not expand on the frequency-domain extraction process here. After obtaining the frequency-domain motion information, we employ a RGB-frequency motion fusion head to integrate motion information from both domains. This fusion head shares the same structure as the long-short motion fusion head, implemented as a two-layer MLP. The fused motion information generated by the RGB-frequency fusion head serves as the key input for updating the prompt.

\noindent \textbf{Process of Online Prompt Updating.} 
After extracting the motion information, we update the target prompt accordingly, enabling our method to adapt to appearance variations induced by motion.
 Specifically, to update the prompt for the current frame (the $k$-th frame) using the target’s motion information $m_k$, we design a blend head $H_b$ that fuses $m_k$ with the prompt of the previous $(k-1)$-th frame. To enhance robustness, the output of the blend head is fused with the prompt from the $(k-1)$-th frame via a residual connection, producing the final prompt for the $k$-th frame. Formally, the prompt updating process for the $k$-th frame is defined as:
\begin{equation}
p_k = (1 - \beta) \cdot H_b\left(p_{k-1}, m_k\right) + \beta \cdot p_{k-1},
\end{equation}
where $\beta$ is a learnable parameter that balances the contributions of the two terms.

The updated prompt $p_k$ is then used in our framework to track the target in the $k$-th frame.

\subsection{Training and Testing Procedures}
\label{subsec:training_and_testing}

After introducing the two components of our method, the \textit{initial prompt learner}, which learns a prompt to activate the target’s bounding box in the first frame through the cross-attention mechanism of the diffusion model, and the \textit{online prompt updater}, which updates the learned prompt based on the target’s motion information to enable continuous tracking, we now provide a detailed description of the training and testing procedures of our approach.

\noindent \textbf{Training.} 
In our method, the trainable network consists of two components. {The first component consists of} the attention map fusion head, embedding projector, and target-shared embedding in the initial prompt learner. The attention map fusion head integrates cross-attention maps from different UNet layers, while the embedding projector adjusts the dimensions of the concatenated target-shared and target-specific embeddings to match the diffusion model's input requirements. For training this first component, we follow USOT~\cite{zheng2021learning} to generate pseudo-labeled training data. We then use the learned prompt from the initial prompt learner to activate corresponding regions in the current frame and compute the loss between these activated regions and the pseudo-labels using~\cref{eq:loss of attention map}. Based on this loss, we update the attention map fusion head, embedding projector, and target-shared embedding through backpropagation.

The second is the Online Prompt Learner, which includes four motion encoders for extracting motion information from video sequences, an RGB-frequency motion fusion head for combining motion information from the RGB and frequency domains, {two long-short motion fusion heads} for merging long-term and short-term motion, and a blend head for updating the prompt with the fused motion information. Note that the image feature extractor in the Online Prompt Learner is a pretrained ResNet-18~\cite{he2016deep} with frozen weights. During training, the motion encoder is initialized with a pretrained ResNet-18 and extended with two Conv3D layers.

\noindent \textbf{Testing.} 
During testing, our method first learns the target-specific embedding $E_{sp}$ through test-time adaptation. Specifically, we use the initial prompt learner to activate the corresponding region in the first frame via our Diff-Tracking framework and compute the loss between this region and the ground-truth bounding box using~\cref{eq:loss of attention map} to update $E_{sp}$. After obtaining $E_{sp}$, we combine it with the target-shared embedding $E_{sh}$ through the embedding projector to generate the initial prompt. This prompt tracks the target from frames 1 to 5. Starting from the sixth frame, the online prompt updater updates the prompt frame-by-frame through the end of the sequence, enabling continuous target tracking.

\section{Experiments}
\label{sec:experiments}
In this section, we first describe the implementation details (\cref{subsec:implementation}), then present main experimental results with comprehensive analysis (\cref{subsec:main results}), followed by ablation studies (\cref{subsec:ablation}), and finally provide visualization results~(\cref{subsec:visual_results}).

\begin{table*}[ht!]
\centering
\caption{We provide experimental results on the TrackingNet~\cite{muller2018trackingnet}, VOT2016~\cite{Kristan2016}, and VOT2018~\cite{Kristan2018} benchmark datasets. Here, ``Unsup'' is an abbreviation for the unsupervised learning. Methods with the “-on” suffix denote the variants that require online updating.}
\label{tab:16-18-trackingnet}
\begin{tabular}{cc|ccc|ccc|ccc}
\toprule[1.2pt] \multirow{2}{*}{ Tracker } & \multirow{2}{*}{ Unsup } & \multicolumn{3}{c|}{ TrackingNet~\cite{muller2018trackingnet} } & \multicolumn{3}{c|}{ VOT~2016~\cite{Kristan2016} } & \multicolumn{3}{c}{ VOT~2018~\cite{Kristan2018} } \\
& & Suc  $\uparrow$ & Pre  $\uparrow$ & NPre  $\uparrow$ & EAO  $\uparrow$ & Acc  $\uparrow$ & Rob  $\downarrow$ & EAO  $\uparrow$ & Acc  $\uparrow$ & Rob  $\downarrow$ \\
\midrule SiamFC~\cite{bertinetto2016fully} & No & 57.1 & 53.3 & 66.3 & 23.5 & 53.2 & 46.1  & 18.8 & 50.3 & 58.5 \\
DaSiamRPN~\cite{zhu2018distractor} & No & -- & -- & -- & 41.1 & 61.0 & 22.0 & 32.6 & 56.0 & 34.0  \\
SiamRPN++~\cite{li2019siamrpn++} & No & 73.3 & 69.4 & 80.0 & -- & -- & -- & 41.4 & 60.0 & 23.4  \\
ATOM~\cite{danelljan2019atom} & No & 70.3 & 64.8 & 71.1 & -- & -- & -- & 40.1 & 59.0 & 20.4  \\
DiMP~\cite{bhat2019learning} & No & 74.0 & 68.7 & 80.1 & -- & -- & -- & 44.0 & 59.7 & 15.3  \\
\midrule KCF~\cite{henriques2014high} & Yes & 44.7 & 41.9 & 54.6 & 19.2 & 48.9 & 56.9 & 13.5 & 44.7 & 77.3  \\
ECO~\cite{danelljan2017eco} & Yes & 56.1 & 48.9 & 62.1 & 37.5 & 55.0 & 56.9 & 28.0 & 27.0 & 48.0 \\
S2SiamFC~\cite{sio2020s2siamfc} & Yes & -- & -- & -- & 21.5 & 49.3 & 63.9 & 18.0 & 46.3 & 78.2 \\
LUDT+~\cite{wang2021unsupervised} & Yes & 56.3 & 49.5 & 63.3 & 29.9 & 57.0 & 33.1 & 23.0 & 49.0 & 41.2 \\
USOT~\cite{zheng2021learning} & Yes & 59.9 & 55.1 & 68.2 & 35.1 & 59.3 & 33.6 & 29.0 & 56.4 & 43.5 \\
USOT-on~\cite{zheng2021learning} & Yes & 61.6 & 56.6 & 69.1 & 40.2 & 60.0 & 23.3 & 34.4 & 57.8 & 30.4 \\
AUDI-T~\cite{kang2025robust} & Yes &61.7&59.4&-&34.1&59.4&33.7&29.8&55.7&42.3\\
 ULAST~\cite{shen2022unsupervised} & Yes & 64.9 & 58.5 & 72.5 & 39.7 & 59.9 & 22.4 & 34.7 & 56.9 & 30.4 \\
ULAST-on~\cite{shen2022unsupervised} & Yes & 65.4 & 59.2 & 73.2 & 41.7 & 60.3 & 21.4 & 35.5 & 57.1 & 28.6 \\
Diff-Tracker~\cite{zhang2024diff} &Yes&67.5&61.4&75.1&43.0&60.5&20.6&36.5&58.0&27.3\\
\midrule
Ours &Yes&70.4&64.1&77.5&45.3&61.1&19.8&38.4&59.4&25.9\\
\bottomrule [1.2pt]
\end{tabular}
\end{table*}

\subsection{Implementation Details}
\label{subsec:implementation}
Below, we first present the pseudo-label generation process for the unsupervised training task, then detail the model architectures of the initial prompt learner and online prompt updater, followed by training and testing details of our method, and finally introduce the evaluation datasets and metrics.

\noindent\textbf{Pseudo-Label Generation Process.}
{To generate pseudo labels for unsupervised training, we follow the established protocol of USOT~\cite{zheng2021learning} and ULAST~\cite{shen2022unsupervised}. Specifically, ARFlow~\cite{liu2020learning} estimates inter-frame optical flow with an adaptive temporal interval. A distance map, computed as the per-pixel L2 deviation from the mean flow, is binarized (threshold $\alpha{=}0.3$) to segment foreground regions. Connected-component analysis on the binary mask yields bounding box candidates, which are filtered by minimum area ($\geq$50\,px), maximum aspect ratio ($\leq$6:1), and scored with a center-biased function. Dynamic programming then selects temporally consistent box sequences using a DIoU-regularized reward ($\gamma{=}4.1$). A video-level quality filter ($Q_v < 0.4$ discarded) and frame-level quality thresholds ($\theta_2{=}0.45$, $\theta_3{=}0.40$) further remove unreliable samples. The filtered pseudo-labels are generated on GOT-10k~\cite{huang2019got}, ImageNet VID~\cite{russakovsky2015imagenet}, LaSOT~\cite{fan2019lasot}, and YouTube-VOS~\cite{xu2018youtube}. For a fair comparison with previous unsupervised trackers~\cite{zheng2021learning,shen2022unsupervised}, we define the ground-truth cross-attention map as the attention map activated only within the bounding box region. Following the analysis in USOT~\cite{zheng2021learning}, the main failure modes of ARFlow + DP include: (i)~\textit{camera shake and occlusion}: these produce noisy flow estimates and hence noisy bounding boxes, which DP and the quality filters are designed to suppress; (ii)~\textit{low-quality videos}: nearly half of all candidate videos are discarded by the video quality threshold.}

\noindent\textbf{Model Architecture.}
Here, we present the model architectures of the initial prompt learner and the online prompt updater.  In the \textit{initial prompt learner}, we employ the pre-trained Stable Diffusion V2.1~\cite{rombach2022high} as our default text-to-image diffusion backbone for the main results, ablation studies, and default efficiency measurements. The embedding projector consists of a two-layer MLP. We set the input image resolution to $512\times512$ and the feature map dimensions ($M_c$, $M_c^\prime$, and $\mathcal{M}$) to $1\times64\times64$. Both target-shared and target-specific embeddings have 1024 dimensions.  The attention map fusion head comprises two fully connected layers with ReLU activation. We extract cross-attention maps from 8 decoder layers. For the \textit{online prompt updater}, we implement both motion encoders by extending a pre-trained ResNet18~\cite{he2016deep} backbone with two Conv3D layers. The image feature extractor is also based on ResNet-18 pre-trained on ImageNet~\cite{deng2009imagenet}. The long-short motion fusion head, RGB-frequency fusion head, and blend head are implemented as two-layer MLPs with ReLU activation. When performing Discrete Cosine Transform (DCT) to transform images from RGB to frequency domain, we follow the standard DCT protocol~\cite{wen2022high} and divide images into $8\times8$ blocks.

\noindent\textbf{Training and Testing Details.}
Our experiments are carried out using RTX 3090 GPUs. During training, we optimize the embedding projector, attention map fusion head, and target-shared embedding $E_{sh}$ in the initial prompt learner using Adam optimizer~\cite{kingma2014adam} with learning rate $5 \times 10^{-4}$ for 20 epochs. For the online prompt updater, we train for 60 epochs using Adam optimizer with learning rate $1 \times 10^{-4}$. {During test-time adaptation, we optimize the target-specific embedding $E_{sp}$ for 3 epochs using Adam optimizer with learning rate $5 \times 10^{-3}$. On an NVIDIA RTX 4090, this one-time initialization takes approximately 1.7 seconds per sequence. After obtaining the initial prompt, it is used to track the target from frames~1 to~5. Starting from frame~6, the online prompt updater refines the prompt at every frame through the end of the sequence based on the target's motion information, enabling continuous adaptation to appearance changes. The online prompt updater refines the prompt and localizes the target at each frame, achieving 35 FPS. For a typical LaSOT sequence, the initialization overhead is amortized to less than 2\% of total inference time.}

\noindent\textbf{Evaluation Datasets and Metrics.} For evaluation, we follow previous work~\cite{shen2022unsupervised,zheng2021learning} and conduct experiments on six challenging tracking datasets: OTB 2015~\cite{7001050}, VOT 2016~\cite{Kristan2016}, VOT 2018~\cite{Kristan2018}, VOT 2020~\cite{kristan2020eighth}, TrackingNet~\cite{muller2018trackingnet}, and LaSOT~\cite{fan2019lasot}. We follow~\cite{shen2022unsupervised} to adopt standard tracking evaluation metrics including Success (Suc), Precision (Pre), Normalized Precision (NPre), Accuracy (Acc), Robustness (Rob), Expected Average Overlap (EAO), and Frames Per Second (FPS). Specifically, Suc measures the area under the curve  of success plots, representing the proportion of frames where the intersection-over-union (IoU)  between predicted and ground-truth bounding boxes exceeds a threshold. Pre computes the percentage of frames with center location error within 20 pixels, while NPre normalizes this error by the ground-truth bounding box size, providing scale-invariant evaluation. Besides, we also report Acc as the average IoU between predictions and ground-truth, Rob as the number of tracking failures requiring re-initialization, and EAO as the primary metric that integrates accuracy and robustness over varying sequence lengths. We additionally report FPS to demonstrate computational efficiency. Finally, following standard tracking benchmark protocols~\cite{paul2022robust}, we report the minimum axis-aligned bounding box enclosing the activated region in the cross-attention map as the tracking result.

\subsection{Main Results}
\label{subsec:main results}

\noindent \textbf{TrackingNet.}
TrackingNet~\cite{muller2018trackingnet} is a large-scale benchmark for evaluating tracking performance in unconstrained environments, containing over 30,000 YouTube video sequences with a test set of 511 videos. The dataset encompasses diverse targets and scenarios, including indoor/outdoor environments, moving objects, pedestrians, vehicles, and animals. Beyond standard precision and success metrics, TrackingNet introduces NPre for scale-invariant evaluation. We evaluate our method against representative supervised methods (\textit{e.g.}, SiamFC~\cite{bertinetto2016fully}, ATOM~\cite{danelljan2019atom}, DiMP~\cite{bhat2019learning}, and OSTrack~\cite{ye2022joint}) and state-of-the-art (SOTA) unsupervised approaches (\textit{e.g.}, LUDT+~\cite{wang2021unsupervised}, USOT~\cite{zheng2021learning},  ULAST~\cite{shen2022unsupervised}, and Diff-Tracker~\cite{zhang2024diff}) on the TrackingNet benchmark. Experimental results (see \cref{tab:16-18-trackingnet}) demonstrate that our method, which exploits semantic and structural knowledge from pre-trained diffusion models, achieves best performance among unsupervised methods. Specifically, our method surpasses the previous best unsupervised method ULAST by 5.4\% in Suc  and achieves performance comparable to supervised methods such as ATOM.

\noindent \textbf{VOT 2016.} 
The VOT 2016 benchmark~\cite{Kristan2016} contains 60 video sequences featuring challenging conditions including drastic scale variations, fast motion, and severe occlusions. Performance is evaluated using three metrics: Rob, Acc, and EAO. As shown in~\cref{tab:16-18-trackingnet}, our method surpasses previous unsupervised approaches, achieving a 2.1\% improvement over the prior SOTA in EAO metric. We attribute the strong performance of our method to the design of the attention harmonization method, which leverages the relationship between target and background in the scene for tracking, thus effectively addressing the challenging tracking scenarios such as drastic scale changes and occlusion present in VOT 2016.

\noindent \textbf{VOT 2018.}
VOT 2018 dataset~\cite{Kristan2018}, similar to VOT 2016, also contains approximately 60 sequences, but with more rigorous annotations and more diverse scenes. Additionally, the testing sequences in VOT 2018 present increased challenges in comparison to those in VOT 2016. The evaluation metrics used to assess the tracker's performance on this dataset are the same as those used in VOT2016. We again conduct comparative experiments (see \cref{tab:16-18-trackingnet}) with previous representative unsupervised methods. The results show that our method achieves significant performance improvements compared to previous methods. For example, in terms of the EAO metric, our method outperforms the previous SOTA unsupervised method ULAST by 3.5\%. We attribute this performance improvement to our method's utilization of the rich knowledge embedded in the diffusion model for tracking tasks. Since the diffusion model has been trained on large amounts of data and possesses strong robustness, our method can effectively perform tracking tasks on the VOT 2018 dataset, which contains more challenging scenarios such as object occlusion.

\begin{table*}[ht!]

\caption{We provide the evaluation results on the OTB2015~\cite{7001050}, LaSOT~\cite{fan2019lasot}, and VOT~2020~\cite{kristan2020eighth} benchmarks. In this table, ``Unsup'' is an abbreviation for the unsupervised learning. Methods with the “-on” suffix denote the variants that require online updating.}
\label{tab:otb15-lasot}
\centering
\begin{tabular}{cc|cc|cc|ccc}
\toprule[1.2pt] \multirow{2}{*}{ Tracker } & \multirow{2}{*}{ Unsup } & \multicolumn{2}{c|}{ OTB2015~\cite{7001050} } & \multicolumn{2}{c|}{ LaSOT~\cite{fan2019lasot} } &\multicolumn{3}{c}{VOT~2020~\cite{kristan2020eighth}} \\
& & Suc  $\uparrow$ & Pre  $\uparrow$ & Suc  $\uparrow$ & Pre  $\uparrow$  & Acc  $\uparrow$ & Rob  $\downarrow$ & EAO  $\uparrow$ \\
\midrule SiamFC~\cite{bertinetto2016fully}  & No & 58.2 & 77.1 & 33.6 & 33.9  & 41.8 & 50.2 &17.9  \\
SiamRPN~\cite{li2018high}  & No & 63.7 & 85.1 & 41.1 & 38.0 &--&--&--\\
SiamRPN++~\cite{li2019siamrpn++}  & No & 69.6 & 92.3 & 49.5 & 49.3&44.2&67.8&24.2 \\
\midrule KCF~\cite{henriques2014high}  & Yes & 48.5 & 69.6 & 17.8 & 16.6& 40.7 &43.2 &15.4 \\
DSST~\cite{danelljan2014accurate}  & Yes & 51.8 & 68.9 & 20.7 & 18.9 &--&--&--\\
LUDT+~\cite{wang2021unsupervised}  & Yes & 63.9 & 84.3 & 30.5 & 28.8&--&--&--  \\
USOT~\cite{zheng2021learning}  & Yes & 58.9 & 80.6 & 33.7 & 32.3 &45.8 &60.0 &22.2 \\
USOT*~\cite{zheng2021learning}  & Yes & 57.4 & 77.5 & 35.8 & 34.0 &44.8 &60.0 &21.9\\
ULAST*-off~\cite{shen2022unsupervised} & Yes & 64.5 & 87.8 & 46.8 & 44.8&--&--&-- \\
ULAST*-on~\cite{shen2022unsupervised} & Yes & 64.8 & 87.9 & 47.1 & 45.1&--&--&-- \\
 Diff-Tracker~\cite{zhang2024diff} &Yes & 66.1 & 89.8 & 48.6 & 47.2 &46.1&58.7&23.9\\
  \midrule
 Ours &Yes &68.7 &91.2 &50.4 &49.2 &47.6&56.8&25.4 \\
\bottomrule[1.2pt]
\end{tabular}
\end{table*}

\noindent \textbf{OTB 2015.}
The OTB 2015 dataset~\cite{7001050} encompasses 100 video sequences and is a classic tracking benchmark, with each sequence having detailed challenge attribute annotations such as illumination changes, occlusion, scale changes, fast motion, blur, and deformation. The evaluation metrics in the OTB2015 benchmark are the success score and precision score. We also compare the performance of our method on the OTB 2015 dataset with representative supervised methods (\textit{e.g.}, SiamFC~\cite{bertinetto2016fully}, SiamRPN~\cite{li2018high}, SiamRPN++~\cite{li2019siamrpn++}) and unsupervised methods (\textit{e.g.}, KCF~\cite{henriques2014high}, DSST~\cite{danelljan2014accurate}, USOT~\cite{zheng2021learning}, ULAST~\cite{shen2022unsupervised}, and Diff-Tracker~\cite{zhang2024diff}). The presented experimental results in~\cref{tab:otb15-lasot} confirm that our method achieves the best performance in terms of robustness, which we attribute to our method inputting the entire frame to our model when learning the prompt and using attention harmonization. These designs allow the prompt to encode the target's semantic, positional, and contextual information.

\noindent \textbf{LaSOT.}
The LaSOT dataset~\cite{fan2019lasot} contains a total of 1,400 sequences (1,120 for training, 280 for testing), with an average of over 2,500 frames per video, making it one of the longest video tracking datasets currently available. The dataset covers 70 categories with balanced category distribution and long-term target presence. It requires trackers to maintain stability and robustness over long time spans while avoiding drift, serving as a key benchmark for evaluating long-term tracking capabilities. The evaluation metric for this benchmark aligns with that of OTB 2015. As can be seen in~\cref{tab:otb15-lasot}, our method surpasses other unsupervised methods on the LaSOT dataset, achieving a performance of {50.4} on the Suc metric. We attribute this superior performance to the inclusion of an Online Prompt Updater in our method, which enables our approach to update the prompt representing the target based on motion information, allowing continuous tracking of the target. This is further confirmed in the ablation study of the online prompt updater in the next subsection.

\noindent \textbf{VOT 2020.}
Compared with VOT 2016~\cite{Kristan2016} and VOT 2018~\cite{Kristan2018}, the VOT 2020 dataset~\cite{kristan2020eighth} poses greater challenges. The video lengths range from several hundred to over one thousand frames, making it significantly longer than VOT 2016 and VOT 2018 datasets. The target categories include humans, animals, vehicles, and common objects, and the scenes span both indoor and outdoor environments. The dataset also contains difficult cases such as heavy occlusion, illumination variations, and motion blur. As reported in \cref{tab:otb15-lasot}, our method achieves the best performance among unsupervised approaches. In terms of the EAO metric, for instance, it outperforms the previous SOTA tracker USOT~\cite{zheng2021learning} by 3.5\%. We attribute this gain to the robust knowledge embedded in the diffusion model, which improves resilience to challenging factors, such as motion blur, in the tracking. This robustness stems from the large-scale datasets used for pretraining of the diffusion model. Moreover, our designed online prompt updater further enhances performance by effectively addressing long video tracking scenarios present in VOT 2020.

\begin{table}[t]
 \caption{We evaluate the effectiveness of the attention harmonization method and the online prompt updater on the VOT 2018~\cite{Kristan2018} benchmark. In the table, ``Ours w/o Attention Harmonization Method'' and ``Ours w/o Online Prompt Updater'' indicate our method with the attention harmonization method removed and with the online prompt updater removed, respectively.}
\label{tab:ablation_1}
\centering
\begin{tabular}{c|ccc}
\toprule[1.2pt]     & EAO $\uparrow$  & Acc $\uparrow$ & Rob $\downarrow$  \\
\midrule 
Ours w/o Attention Harmonization Method &  37.1 & 58.3 & 27.8 \\
Ours w/o Online Prompt Updater &    35.5 & 57.6 & 28.5\\
\midrule 
Ours  &38.4 &59.4 &25.9  \\
\bottomrule [1.2pt]
\end{tabular}
\end{table}

\subsection{Ablation Study}
\label{subsec:ablation}

\noindent \textbf{Attention Harmonization Method.}
We also investigate the effect of harmonizing the self-attention map and the cross-attention map (see ``Ours w/o Attention Harmonization Method'' in \cref{tab:ablation_1}). Without the attention harmonization method, the loss of our method (\cref{eq:loss of attention map}) is computed directly between the cross-attention map $M_c$, extracted from the UNet of the diffusion model, and the generated ground-truth cross-attention map. As shown in \cref{tab:ablation_1}, incorporating attention harmonization leads to improved performance. This improvement arises because attention harmonization enables the model to exploit the relationship between the target and its background for more effective tracking.

\noindent \textbf{Online Prompt  Updater.}
As shown in \cref{tab:ablation_1}, we further evaluate the effect of the online prompt updater in our framework. Without this component, our Diff-Tracking method relies solely on the learned initial prompt for tracking in subsequent frames. 
The experimental results demonstrate that incorporating the online prompt updater improves the performance of our method. This improvement stems from its ability to update the target-representative prompt using the object’s motion information, which enables more effective tracking in long video sequences.

\noindent \textbf{Target-Shared Embedding $E_{sh}$.} We introduce the target-shared embedding $E_{sh}$ to capture features that are shared across tracking targets, such as the common property that most tracked objects are salient, which enhances the robustness of our method. We validate this design on the challenging LaSOT~\cite{fan2019lasot} dataset. When $E_{sh}$ is removed, the Embedding Projector is also discarded. In this setting, only the Attention Map Fusion Head is trained during the stage where $E_{sh}$ would otherwise be learned. As indicated by the Suc metric in the experimental results of \cref{tab:ablation_2}, incorporating $E_{sh}$ improves robustness and yields performance gains.

\begin{table}[t]

 \caption{We evaluate the effectiveness of the proposed Target-Shared Embedding $E_{sh}$, the Attention Map Fusion Head that aggregates cross-attention maps from different UNet layers, and the use of Frequency-Domain Motion Information on the challenging long-term tracking benchmark LaSOT~\cite{fan2019lasot}..}
\label{tab:ablation_2}
\centering
\begin{tabular}{c|cc}

\toprule [1.2pt]      & Suc  $\uparrow$ & Pre  $\uparrow$  \\
\midrule 
Ours w/o Target-Shared Embedding $E_{sh}$  &49.1  & 48.9   \\
  Ours w/o Attention Map Fusion Head  & 49.5 & 48.3    \\
    Ours w/o Frequency-Domain Motion Information  & 49.2 & 48.3    \\
\midrule 
Ours &50.4 &49.2   \\
\bottomrule [1.2pt]
\end{tabular}

\end{table}

\noindent \textbf{Attention Map Fusion Head.}
In this paper, we design an Attention Map Fusion Head to leverage the cross-attention maps from different layers of the diffusion model’s UNet, as they capture different types of features. For example, intermediate layers emphasize semantic structure, while deeper layers focus on fine-grained textures~\cite{zhang2023tale}.  When the Attention Map Fusion Head is removed, we follow \cite{zhang2024diff} and use only a single cross-attention map. As demonstrated in the experimental results (see \cref{tab:ablation_2}), incorporating the Attention Map Fusion Head to utilize these complementary features leads to additional performance improvements for our method.

\begin{figure*}[t]
\centering
\includegraphics[width=7.25in]{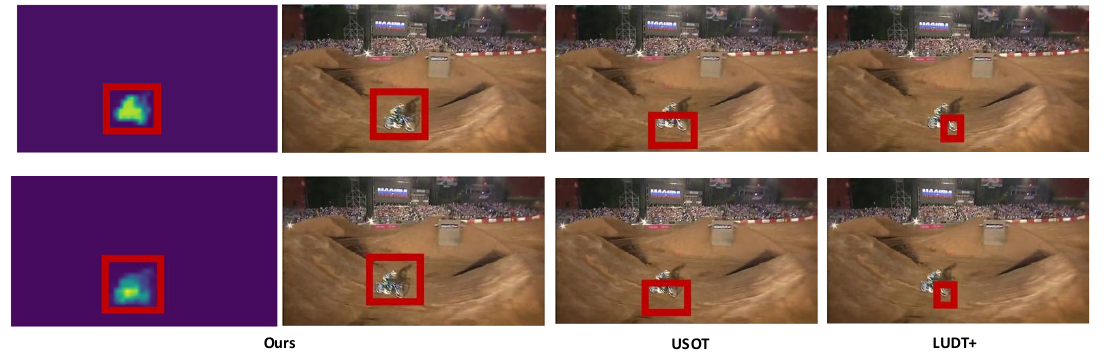}
\caption{We provide visualization results on a long and challenging video from the VOT 2018 benchmark~\cite{Kristan2018}, comparing our method with representative unsupervised trackers USOT~\cite{zheng2021learning} and LUDT+~\cite{wang2021unsupervised}. The visualizations of our method (``Ours in the figure") include both the activated regions on the cross-attention map and the corresponding tracking outputs.}
\label{fig:vis_2}
\end{figure*}

\noindent \textbf{Frequency-Domain Motion Information.}
To handle tracking in scenarios where the target and background share similar colors, we introduce the frequency-domain motion information. Although the target and background may look alike in color, they often differ in texture~\cite{ning2009robust,chase}, and frequency representations are effective in capturing these differences~\cite{hu2024sfdfusion}. This enables the extraction of more robust motion information in the frequency domain under such challenging conditions, thereby improving the robustness of our method. When Frequency-Domain Motion Information is removed, only the RGB-domain input is kept in the online prompt updater, and the RGB-Frequency Fusion Head is discarded. We validate this design on LaSOT~\cite{fan2019lasot}, and as shown in {\cref{tab:ablation_2}}, removing Frequency-Domain Motion Information results in a 0.8\% decrease in Pre, confirming its effectiveness.

\begin{table}[t]

 \caption{We assess the impact of long-term and short-term motion information on the challenging long-term tracking benchmark LaSOT~\cite{fan2019lasot}.}
\label{tab:ablation_3}
\centering
\begin{tabular}{c|cc}

\toprule [1.2pt]      & Suc  $\uparrow$ & Pre  $\uparrow$  \\
\midrule 
Ours w/o Long-Term Motion Information  &47.1  & 46.2   \\
  Ours w/o Short-Term Motion Information  & 48.4 & 47.5    \\
\midrule 
Ours &50.4 &49.2   \\
\bottomrule [1.2pt]
\end{tabular}

\end{table}

\noindent \textbf{Long-Term and Short-Term Motion Information.}
To further examine the effect of different types of motion information (long-term and short-term) on our method, we perform ablation studies on the LaSOT~\cite{fan2019lasot} dataset. When one type of motion information is removed, its corresponding motion encoder and the Long-Short Motion Fusion Head in the online prompt updater are also discarded. As shown in \cref{tab:ablation_3}, incorporating short-term motion information for online updating yields performance gains, since short-term motion captures the target’s immediate movement and provides timely cues for prompt updating. Furthermore, integrating long-term motion information leads to even larger improvements, which can be attributed to the stronger spatio-temporal consistency offered by long-term motion compared with short-term motion.

\subsection{More Experiments}
\label{subsec:visual_results}

\noindent \textbf{Visual Results.}
We present visualization results comparing our method with SOTA unsupervised trackers (USOT~\cite{zheng2021learning} and LUDT+~\cite{wang2021unsupervised}) on a long and challenging video from the VOT 2018 benchmark~\cite{Kristan2018}. As shown in \cref{fig:vis_2}, our method accurately activates the target regions on the attention map while suppressing the background, highlighting the precision of our approach. In particular, compared with SOTA methods (USOT and LUDT+), our method achieves higher accuracy by activating both the rider and the motorcycle regions. By contrast, USOT and LUDT+ generate bounding boxes that cover only the motorcycle and the motorcycle’s front wheel,~respectively.

{
\noindent \textbf{Qualitative Analysis of $E_{sh}$.} To qualitatively analyze the role of $E_{sh}$, we visualize the cross-attention maps of the full approach and the variant without $E_{sh}$ on two hard cases from VOT 2018~\cite{Kristan2018} (\cref{fig:esh_vis}). In the first example, a cyclist is severely occluded by surrounding trees; relying on $E_{sp}$ alone (i.e., removing $E_{sh}$) causes the attention response to nearly vanish, as the target-specific semantics alone cannot resolve the target under heavy occlusion. In the second example, camera shake introduces noticeable motion blur; without $E_{sh}$, the attention map shrinks and fails to cover the full extent of the target. In both cases, the full approach with $E_{sh}$ produces well-focused attention maps that accurately capture the target, confirming that $E_{sh}$ provides a complementary generic saliency prior that stabilizes the attention when $E_{sp}$ alone is insufficient.

{
\noindent \textbf{Semantic Interpretability.} To examine whether the learned embeddings retain semantic interpretability, we compute the cosine similarity between each learned embedding and the word embeddings of common English words encoded by the CLIP text encoder~\cite{radford2021learning}. As shown in \cref{tab:nn_semantics}, on the \textit{motorcycle} sequence from \cref{fig:vis_2} where a person rides a motorcycle, $E_{sp}$'s top neighbors include both ``motorcycle'' and ``rider'', indicating that $E_{sp}$ captures the composite visual concept rather than a single noun label. Meanwhile, $E_{sh}$'s nearest neighbors are generic foreground-related words (``object'', ``foreground'', ``thing''), confirming its role as a class-agnostic saliency prior. These results verify that both embeddings operate within the diffusion model's meaningful semantic space rather than collapsing into meaningless noise, supporting our claim that the method genuinely leverages the semantic knowledge encoded in the pretrained diffusion~model.

\begin{figure}[t]
\centering
\includegraphics[width=\linewidth]{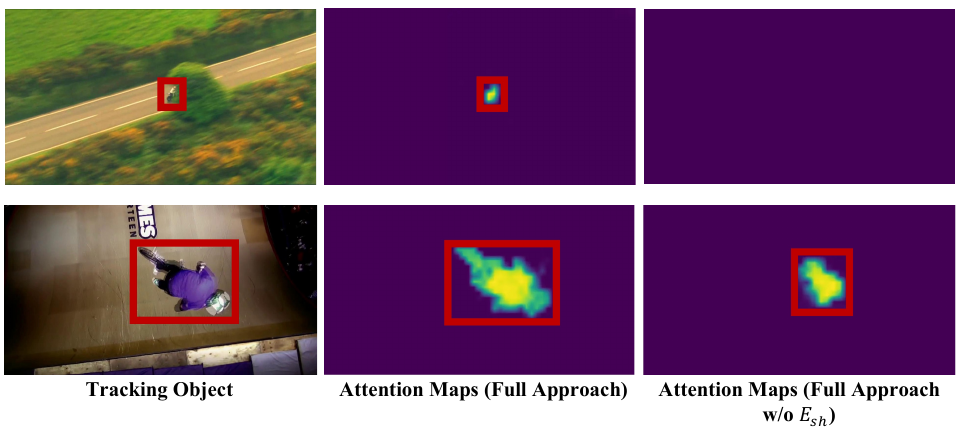}
\caption{Cross-attention map visualization on two hard cases from VOT 2018~\cite{Kristan2018}. From left to right: input frame with ground-truth bounding box, attention map of the full approach, and attention map without $E_{sh}$. Top: a cyclist under severe occlusion from trees---removing $E_{sh}$ causes the attention to nearly vanish. Bottom: a skateboarder with noticeable motion blur due to camera shake---removing $E_{sh}$ yields an incomplete activation that fails to cover the full target extent.}
\label{fig:esh_vis}
\end{figure}
}

\begin{table}[t]
\caption{Nearest-neighbor analysis of learned embeddings in the CLIP vocabulary space on the \textit{motorcycle} sequence. $E_{sp}$ captures target-specific semantics; $E_{sh}$ captures class-agnostic foreground saliency. Values in parentheses are cosine similarities.}
\label{tab:nn_semantics}
\centering
\begin{tabular}{c|l}
\toprule[1.2pt]
\textbf{Embedding} & \textbf{Nearest Words (Cosine Sim.)} \\
\midrule
$E_{sp}$ & motorcycle (.80), rider (.73), bike (.71) \\
\midrule
$E_{sh}$ & object (.71), foreground (.68), thing (.65) \\
\bottomrule[1.2pt]
\end{tabular}
\end{table}
}

\begin{table}[t]

 \caption{We report the performance and inference speed of our method in comparison with state-of-the-art unsupervised trackers on the LaSOT~\cite{fan2019lasot} dataset.}
\label{tab:inference}
\centering
\begin{tabular}{c|ccc}

\toprule [1.2pt]      & Suc  $\uparrow$ & Pre  $\uparrow$ & FPS $\uparrow$ \\
\midrule 
LUDT+~\cite{zheng2021learning}   &30.5  & 28.8 & 46  \\
  USOT~\cite{wang2021unsupervised}   & 33.7 & 32.3 & 31   \\
    Diff-Tracker~\cite{zhang2024diff}   & 48.6 & 47.2 & 36   \\
\midrule 
Ours &50.4 &49.2 & 35   \\
\bottomrule [1.2pt]
\end{tabular}

\end{table}

\begin{table}[t]

 \caption{
To assess the generalization ability of our method, we evaluate its performance on the LaSOT~\cite{fan2019lasot} dataset with different diffusion model backbones. In the table, ``Ours'' denotes the default version that uses Stable Diffusion V2.1~\cite{rombach2022high} as the backbone.
}
\label{tab:backbones}
\centering
\begin{tabular}{c|ccc}

\toprule [1.2pt]      & Suc  $\uparrow$ & Pre  $\uparrow$ & FPS $\uparrow$ \\
\midrule 
Ours w/ Pixart \cite{chen2023pixart}   &50.5  & 49.3 & 21  \\
Ours w/ FLUX \cite{flux2024}   & 50.7 & 49.5 & 8   \\
\midrule 
Ours &50.4 &49.2 & 35   \\
\bottomrule [1.2pt]
\end{tabular}

\end{table}

\noindent \textbf{Inference Speed.}
We present the performance of our method and SOTA unsupervised trackers (USOT~\cite{zheng2021learning}, LUDT+~\cite{wang2021unsupervised}, and Diff-Tracking~\cite{zhang2024diff}) on the LaSOT~\cite{fan2019lasot} dataset, along with their inference speeds measured on an RTX 4090 GPU. As shown in \cref{tab:inference}, while our method is not the fastest, it operates in real time and achieves a favorable trade-off between accuracy and speed. More importantly, it delivers substantially higher accuracy than previous methods. For instance, in terms of the Suc, {our method attains} {50.4} compared with 30.5 for LUDT+.

{
\noindent \textbf{Camouflage Analysis.} To verify the generalizability of the proposed RGB-to-Frequency Transformation (abbreviated as ``Freq'' in \cref{tab:freq_camo}), we integrate it into SimTrack~\cite{chen2022simtrack}, a supervised tracker, and evaluate on the COTD~\cite{guo2024cotd} camouflaged object tracking benchmark. As shown in \cref{tab:freq_camo}, adding the frequency-domain transformation improves SimTrack's AUC from 63.9 to \textbf{65.5} (+1.6), demonstrating that the frequency-domain representation is a general and transferable module that benefits trackers in camouflage-like scenarios. This improvement arises because targets and backgrounds that appear similar in the RGB domain often exhibit distinct frequency-domain signatures due to texture differences, enabling more effective target-background discrimination.

\begin{table}[t]
\caption{Effect of integrating our RGB-to-Frequency Transformation (``Freq'') into SimTrack~\cite{chen2022simtrack} on the COTD~\cite{guo2024cotd} camouflaged object tracking benchmark (AUC score).}
\label{tab:freq_camo}
\centering
\begin{tabular}{l|c}
\toprule[1.2pt]
 & AUC $\uparrow$ \\
\midrule
SimTrack~\cite{chen2022simtrack} & 63.9 \\
SimTrack + Freq (Ours) & \textbf{65.5} \\
\bottomrule[1.2pt]
\end{tabular}
\end{table}
}

{
\noindent \textbf{Efficiency Analysis.} As Table~\ref{tab:inference} shows, the FPS difference between our method and the conference version (Diff-Tracker) is marginal (35 vs.\ 36 FPS), confirming that the diffusion U-Net is the computational bottleneck. To explore practical pathways for inference optimization, we replace the default {Stable Diffusion v2.1}~\cite{rombach2022high} backbone with BK-SDM~\cite{kim2024bksdm}, a knowledge-distilled variant of SD that compresses the U-Net architecture. As shown in \cref{tab:efficiency}, BK-SDM reduces the parameter count by 27\% and GPU memory by 24\%, while achieving comparable tracking accuracy (50.7 vs.\ 50.4 Suc on LaSOT). On an NVIDIA RTX 4090, the distilled backbone runs at 49 FPS (a 40\% speedup). To simulate edge-application scenarios, we also benchmark on a consumer-grade RTX 3060 (12 GB), where it achieves 25 FPS with only 5.1 GB memory, demonstrating the feasibility of deploying our framework on lower-end hardware.

\begin{table}[t]
\caption{Efficiency comparison on LaSOT~\cite{fan2019lasot} between {SD v2.1}~\cite{rombach2022high} and the distilled BK-SDM~\cite{kim2024bksdm}. ``Mem'' = peak GPU memory.}
\label{tab:efficiency}
\centering
\begin{tabular}{l|ccccc}
\toprule[1.2pt]
 & Mem & \multicolumn{2}{c}{FPS $\uparrow$} & Suc $\uparrow$ & Pre $\uparrow$ \\
\cmidrule(lr){3-4}
 & (GB) & 4090 & 3060 & & \\
\midrule
{SD v2.1}~\cite{rombach2022high} & 6.7 & 35 & 17 & 50.4 & 49.2 \\
BK-SDM~\cite{kim2024bksdm} & 5.1 & 49 & 25 & \textbf{50.7} & \textbf{49.3} \\
\bottomrule[1.2pt]
\end{tabular}
\end{table}
}

\noindent \textbf{Generalization Ability.} To further assess the generalization ability of our approach, we also evaluate it on the LaSOT~\cite{fan2019lasot} dataset with more advanced diffusion models (\textit{e.g.}, Pixart~\cite{chen2023pixart} and FLUX~\cite{flux2024}) as backbones. As shown in \cref{tab:backbones}, our method maintains strong performance with these alternative diffusion models, highlighting its generalization capability. While more advanced diffusion models bring performance improvements, their larger parameter sizes reduce real-time efficiency. Our method achieves a favorable trade-off between accuracy and efficiency.

{
\noindent \textbf{Contribution of Large-Scale Pretrained Backbone.} To isolate the contribution of the pretrained Stable Diffusion backbone from our proposed designs, we conduct two sets of experiments. First, we construct a Naive SD Baseline that uses the same frozen Stable Diffusion cross-attention maps but removes all proposed components (attention harmonization, attention map fusion head, target-shared/specific embeddings, online prompt updater, and frequency-domain branch) and instead uses the ground-truth category name of the tracking target as the text prompt for localization. Second, we replace the CNN feature backbones of two open-source unsupervised trackers, USOT~\cite{zheng2021learning} and LUDT+~\cite{wang2021unsupervised}, with the same Stable Diffusion intermediate features used by our method. As shown in \cref{tab:contribution_analysis}, the Naive SD Baseline achieves only 17.1 EAO on VOT 2018 (far below our 38.4 EAO), and equipping existing trackers with SD features yields only marginal gains, confirming that our performance improvements stem from the proposed tracking-specific designs rather than from large-scale pretraining alone.

\begin{table}[t]
\caption{Contribution analysis on VOT 2018~\cite{Kristan2018} isolating the effect of pretraining from our proposed designs. All SD-based variants share the same frozen Stable Diffusion~\cite{rombach2022high}.}
\label{tab:contribution_analysis}
\centering
\begin{tabular}{l|ccc}
\toprule[1.2pt]
 & EAO $\uparrow$ & Acc $\uparrow$ & Rob $\downarrow$ \\
\midrule
Naive SD Baseline & 17.1 & 45.6 & 58.3 \\
\midrule
LUDT+~\cite{wang2021unsupervised} (original) & 23.0 & 49.0 & 41.2 \\
LUDT+~\cite{wang2021unsupervised} w/ SD features & 25.3 & 50.4 & 38.2 \\
\midrule
USOT~\cite{zheng2021learning} (original) & 29.0 & 56.4 & 43.5 \\
USOT~\cite{zheng2021learning} w/ SD features & 31.2 & 57.8 & 41.3 \\
\midrule
\textbf{Ours} & \textbf{38.4} & \textbf{59.4} & \textbf{25.9} \\
\bottomrule[1.2pt]
\end{tabular}
\end{table}
}

{
\noindent \textbf{Pseudo-Label Quality Sensitivity.} To investigate the sensitivity of our method to pseudo-label quality, we train the proposed tracker using pseudo-labels generated by three optical flow models of varying quality: TV-L1~\cite{zach2007tvl1}, ARFlow~\cite{liu2020learning}, and SEA-RAFT~\cite{wang2024sea}. As reported in \cref{tab:flow_sensitivity}, the performance variation across these configurations is remarkably small ($\leq$1.0\% in Success), suggesting that our method exhibits robustness to pseudo-labels of varying quality during training. We conjecture that the frozen pretrained diffusion features, which already encode rich semantic priors, may help mitigate the impact of noisy supervision to some extent.

\begin{table}[t]
\caption{Pseudo-label quality sensitivity on LaSOT~\cite{fan2019lasot}. Three optical flow models of increasing quality are used to generate training labels. The gap is $\leq$1.0\% Suc, confirming robustness to label noise.}
\label{tab:flow_sensitivity}
\centering
\begin{tabular}{l|cc}
\toprule[1.2pt]
\textbf{Flow Model} & \textbf{Suc} $\uparrow$ & \textbf{Pre} $\uparrow$ \\
\midrule
TV-L1~\cite{zach2007tvl1}       & 49.8 & 48.7 \\
ARFlow~\cite{liu2020learning}    & 50.4 & 49.2 \\
SEA-RAFT~\cite{wang2024sea}      & \textbf{50.8} & \textbf{49.5} \\
\bottomrule[1.2pt]
\end{tabular}
\end{table}

To further analyze the robustness of our method to pseudo-label noise, we vary the video-level quality threshold $Q_v$ used to filter the training set. Lowering $Q_v$ from the default 0.4 to 0.3 and 0.2 progressively includes more low-quality videos with noisier pseudo-labels. As reported in \cref{tab:flow_sensitivity_qv}, even with a relaxed threshold of 0.2, the performance drops by only $-$0.6 Suc and $-$0.6 Pre. Specifically, reducing $Q_v$ from 0.4 to 0.3 introduces a marginal $-$0.4 Suc degradation, while further lowering it to 0.2 adds only an additional $-$0.2 Suc. The graceful degradation across both thresholds suggests that our method does not heavily rely on high-quality pseudo-labels. We attribute this robustness to the frozen pretrained diffusion model, whose cross-attention maps already encode strong priors about object boundaries and spatial structure, allowing the learned embeddings to remain effective even when trained with noisier supervision.

\begin{table}[t]
\caption{Robustness to pseudo-label noise on LaSOT~\cite{fan2019lasot}. We vary the video-level quality threshold $Q_v$ to include progressively noisier training videos. Lower thresholds admit more low-quality pseudo-labels.}
\label{tab:flow_sensitivity_qv}
\centering
\begin{tabular}{c|cc}
\toprule[1.2pt]
$Q_v$ \textbf{Threshold} & \textbf{Suc} $\uparrow$ & \textbf{Pre} $\uparrow$ \\
\midrule
\textbf{0.4} & \textbf{50.4} & \textbf{49.2} \\
0.3 & 50.0 & 48.9 \\
0.2 & 49.8 & 48.6 \\
\bottomrule[1.2pt]
\end{tabular}
\end{table}
}

{
\begin{table}[t!]
\caption{Effect of online prompt updater starting frame on LaSOT~\cite{fan2019lasot}.}
\label{tab:start_frame}
\centering
\begin{tabular}{c|cc}
\toprule[1.2pt]
\textbf{Start Frame} & \textbf{Suc} $\uparrow$ & \textbf{Pre} $\uparrow$ \\
\midrule
2 & 49.7 & 48.5 \\
4 & 50.1 & 48.8 \\
\textbf{6} & \textbf{50.4} & \textbf{49.2} \\
8 & 50.2 & 49.1 \\
10 & 49.8 & 48.7 \\
\bottomrule[1.2pt]
\end{tabular}
\end{table}
}

{
\noindent \textbf{Starting Frame of Online Prompt Updater.} We conduct an ablation study on the starting frame of the online prompt updater. As shown in \cref{tab:start_frame}, starting at frame~6 yields the best performance. In the first few frames of a video sequence, the target's appearance generally remains close to its initial state, so the initial prompt learned from the first frame provides sufficient discriminative information for accurate tracking without online updates. Starting the updater too early (e.g., frame~2 or~4) introduces unnecessary prompt modifications during this stable period, which can add noise rather than useful adaptation signals. On the other hand, delaying the start beyond frame~6 (e.g., frame~8 or~10) means the prompt remains static while the target's appearance has already begun to change, leading to slightly lower accuracy. We therefore adopt frame~6 as the default starting point.
}

{
\noindent \textbf{Update Interval Ablation.} To analyze how the prompt update frequency affects tracking performance, we vary the interval between consecutive updates on LaSOT~\cite{fan2019lasot}. As shown in \cref{tab:update_interval}, updating the prompt at every frame (interval\,=\,1) achieves the best accuracy (\textbf{50.4} Suc). Increasing the interval to 3 or 5 frames causes gradual accuracy degradation ($-$0.9 and $-$1.2 Suc, respectively), as the prompt becomes less responsive to appearance changes. At interval\,=\,10, accuracy drops by $-$1.8 Suc. Notably, the FPS remains largely unchanged (35--38) because the U-Net forward pass dominates per-frame cost and the online prompt updater itself is lightweight (motion encoders + MLP fusion heads). We therefore adopt per-frame updating as the default for the best~accuracy.

\begin{table}[t]
\caption{Effect of prompt update interval on LaSOT~\cite{fan2019lasot}. ``Interval'' = number of frames between consecutive online prompt updates.}
\label{tab:update_interval}
\centering
\begin{tabular}{c|ccc}
\toprule[1.2pt]
\textbf{Interval} & \textbf{Suc} $\uparrow$ & \textbf{Pre} $\uparrow$ & \textbf{FPS} $\uparrow$ \\
\midrule
\textbf{1} & \textbf{50.4} & \textbf{49.2} & 35 \\
3 & 49.5 & 48.4 & 36 \\
5 & 49.2 & 47.7 & 36 \\
10 & 48.6 & 47.1 & 38 \\
\bottomrule[1.2pt]
\end{tabular}
\end{table}
}

{
\begin{table}[h]
\caption{Component-wise timing analysis on VOT 2018~\cite{Kristan2018}. ``Init'' = time spent by the Initial Prompt Learner; ``s/seq'' = average seconds per sequence.}
\label{tab:timing_breakdown}
\centering
\begin{tabular}{l|ccc}
\toprule[1.2pt]
\textbf{Configuration} & \textbf{Init} (s) & \textbf{FPS} & \textbf{s/seq} \\
\midrule
Full method & 1.7 & 35 & 11.7 \\
w/o Initial Prompt Learner & 0 & 36 & 10.1 \\
w/o Online Prompt Updater & 1.7 & 38 & 9.3 \\
\bottomrule[1.2pt]
\end{tabular}
\end{table}
}

{
\noindent \textbf{Component Timing Analysis.} \cref{tab:timing_breakdown} reports the per-component timing overhead on VOT 2018. Removing the initial prompt learner eliminates the 1.7\,s initialization cost ($-$1.6\,s/seq). Removing the online prompt updater increases FPS from 35 to 38 ($-$2.4\,s/seq). Both components add minimal overhead while delivering the accuracy gains reported in~\cref{tab:ablation_1}.
}

\section{Conclusion}
\label{sub:conclusion}
In this paper, we address the challenging problem of unsupervised visual tracking from a new perspective by introducing Diff-Tracking, which exploits the rich knowledge of text-to-image diffusion models through their cross-attention mechanism. Diff-Tracking first learns an initial prompt to represent the target using an initial prompt learner, and then updates the prompt based on target motion with an online prompt updater. Extensive experiments show that Diff-Tracking surpasses state-of-the-art unsupervised trackers on six widely used benchmarks. We also conduct detailed ablation studies to verify the effectiveness of each proposed component. {Furthermore, our contribution analysis confirms that the performance gains are driven by our tracking-specific designs rather than by the pretrained Stable Diffusion~\cite{rombach2022high} alone, as evidenced by the substantial performance gap between our full method and both a Naive SD Baseline and existing trackers equipped with SD features.}

\section*{Acknowledgment}
This work was supported by the NSFC Regional Innovation and Development Joint Fund under Grant U25A20537, the National Key Research and Development Program of China No. 2024YFC3015600.

\bibliographystyle{IEEEtran}
\bibliography{IEEEabrv,main}

\vfill

\end{document}